\documentclass{article} 
\usepackage{iclr2027_conference,times}

\usepackage{amsmath,amsfonts,bm}

\def\eqref#1{equation~\ref{#1}}

\def\1{\bm{1}}

\DeclareMathAlphabet{\mathsfit}{\encodingdefault}{\sfdefault}{m}{sl}
\SetMathAlphabet{\mathsfit}{bold}{\encodingdefault}{\sfdefault}{bx}{n}

\usepackage{hyperref}
\usepackage{url}
\usepackage{xspace}

\usepackage{graphicx}  
\usepackage{booktabs}  
\usepackage{multirow}  

\usepackage{xcolor}

\newcommand{\modelname}{\textsc{\textbf{StreamingSwap}}\xspace}
\newcommand{\lossname}{\textit{Joint Rec-ID Loss}\xspace}
\newcommand{\roiname}{\textit{Face-RoI Generation}\xspace}
\newcommand{\contextname}{\textit{Short-Context Rollout Training}\xspace}
\newcommand{\datasetname}{\textbf{HumanFaceVid}\xspace}

\AddToHook{env/itemize/before}{%
    \setlength{\leftmargini}{1.2em}%
}

\title{From Short Clips to Real-Time Streams: Identity-Aware Supervision and Context-Efficient Generation for Video Face Swapping}

\author{
\makebox[\dimexpr\textwidth-2\tabcolsep\relax][c]{%
  \textbf{Zekai Luo}$^{1}$\thanks{Equal contribution.}
  \quad \textbf{Yuwei Liang}$^{1}$\footnotemark[1]
  \quad \textbf{Ruiqi Ma}$^{1}$
  \quad \textbf{Haoyu Huang}$^{1}$
  \quad \textbf{Xuehao Wang}$^{1}$
  \quad \textbf{Huanyi Zheng}$^{1}$%
}
\\[3pt]
\makebox[\dimexpr\textwidth-2\tabcolsep\relax][c]{%
  \textbf{Mingyu Liu}$^{1}$
  \quad \textbf{Linhao Zhong}$^{1}$
  \quad \textbf{Chenchen Jing}$^{2}$
  \quad \textbf{Yuling Xi}$^{1}$\thanks{Corresponding authors.}
  \quad \textbf{Hao Chen}$^{1}$\footnotemark[2]%
}
\\[4pt]
\makebox[\dimexpr\textwidth-2\tabcolsep\relax][c]{%
  \normalfont
  $^{1}$Zhejiang University
  \quad
  $^{2}$Zhejiang University of Technology%
}
\\[8pt]
\makebox[\dimexpr\textwidth-2\tabcolsep\relax][c]{%
  \normalfont
  \href{https://aim-uofa.github.io/StreamingSwap/}
       {\texttt{https://aim-uofa.github.io/StreamingSwap/}}%
}
}

\iclrfinalcopy 
\begin{document}

\maketitle

\fancyhead{}
\renewcommand{\headrulewidth}{0pt}

\begin{figure*}[h]
    \centering
    \vspace{-15pt}
    \includegraphics[width=1.\linewidth]{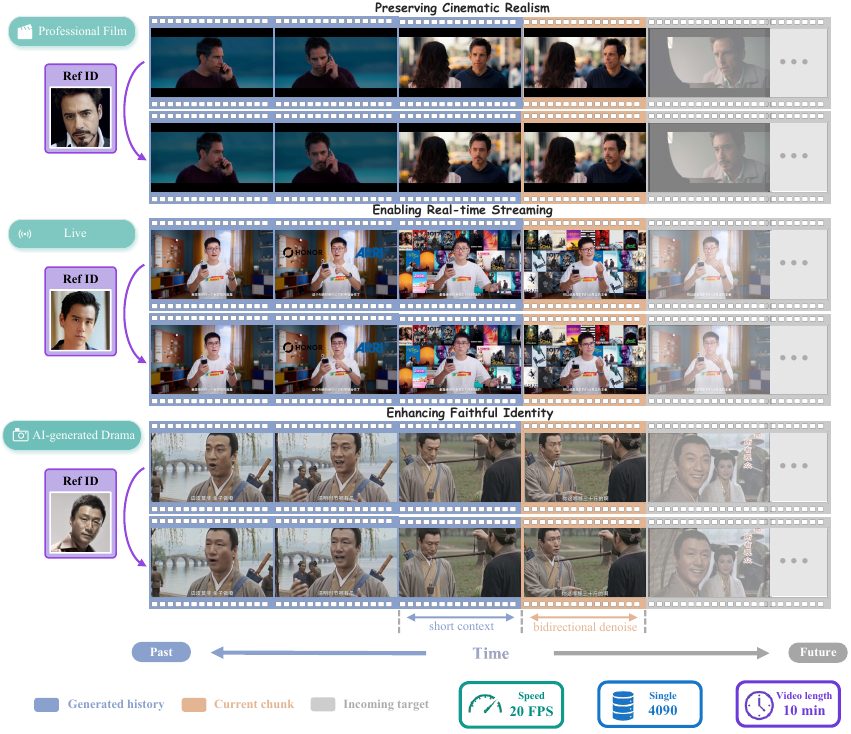}
    \vspace{-20pt}
    \caption{
        \textbf{Real-time streaming video face swapping with \modelname.}
        Examples span film, live, and AI-generated drama, with reference videos above and swapped outputs below.
    }
    \label{fig:teaser}
\end{figure*}

\begin{abstract}

Video face swapping transfers the reference identity to a target video while preserving its other attributes and motion.
Despite recent progress in reference-guided flow-based methods, jointly achieving faithful identity transfer, high fidelity, and real-time generation over long video streams remains challenging.
To address this challenge, we rethink the task-specific structure of reference-guided video face swapping.
First, face swapping changes only identity while preserving most visual content, yet flow-matching supervision is weakly sensitive to these changes.
We therefore develop \lossname, which balances explicit identity supervision and latent reconstruction according to the noise level.
Second, the target video already provides non-edited attributes and temporal dynamics, enabling generation with fewer spatial tokens and shorter temporal contexts through \roiname and \contextname.
By integrating these designs, we propose \modelname, a reference-guided flow-matching framework for real-time streaming video face swapping.
We also introduce \datasetname, comprising 1,000 hours of cinematic training data and evaluation sets of 200 short and 200 long videos.
Experiments demonstrate that our method achieves competitive identity transfer and visual quality at 20 FPS on a single NVIDIA RTX 4090 GPU, with stable generation over videos up to 10 minutes long.

\end{abstract}    
\section{Introduction}
\label{sec:intro}

Video face swapping transfers identity from a source image to a target video while preserving other target attributes, such as expression, pose, gaze, illumination, and background.
It has broad applications in film production, AI-generated dramas, and live streaming.
GAN-based approaches~\citep{chen2020simswap,shiohara2023blendface,luo2025canonswap,facefusion2025} offer strong identity transfer and efficient inference, but processing frames independently can introduce visual artifacts and temporal flicker.
To improve realism and temporal coherence, diffusion-based approaches leverage the generative priors of pretrained video diffusion models.
Among these, inpainting-based approaches~\citep{han2024face,wang2025dynamicface,chen2024hifivfs,shao2024vividface} mask the target face and reconstruct it conditioned on source identity and auxiliary attributes, but removing direct access to fine-grained target cues makes non-edited attributes difficult to preserve faithfully.



Recent reference-guided methods, represented by LivingSwap~\citep{luo2026preserving} and DreamID-V~\citep{guo2026dreamid}, construct paired supervision to retain the complete target video as a visual reference, enabling more faithful preservation of fine-grained target attributes.
However, the unmasked target face contains not only the attributes to be preserved but also the identity to be replaced, creating a competing identity signal that makes source-identity injection difficult.
Meanwhile, these methods are fine-tuned from large-scale video diffusion models~\citep{wan2025wan} and remain computationally expensive due to iterative full-frame denoising over long temporal contexts.
Consequently, extending their high-fidelity results to real-time, long-duration generation while ensuring faithful source-identity transfer remains difficult.

These challenges motivate us to rethink reference-guided video face swapping through its task-specific structure.
First, video face swapping changes only identity while preserving most visual content in the target video.
This leaves flow-matching supervision with a weak signal for identity transfer, as identity differences account for only a small fraction of the latent reconstruction error.
Second, the target video directly provides the non-edited attributes and temporal dynamics to be preserved.
Yet existing reference-guided methods still process background over long temporal contexts, resulting in substantial computational redundancy.



Guided by these observations, we introduce \modelname, the first reference-guided flow-matching framework for real-time, long-duration video face swapping.
To strengthen source-identity transfer, \lossname combines clean-latent reconstruction with explicit identity supervision, balancing their relative contributions according to the noise level.
For efficient generation, \roiname restricts video conditioning and generation to face-region tokens, avoiding redundant background computation.
To enable streaming, \contextname trains the bidirectional model through autoregressive rollout with a short generated history, reducing the mismatch between training and inference while retaining bidirectional generation within each window.
We further introduce \datasetname, a cinematic video face-swapping dataset constructed from OpenHumanVid~\citep{li2025openhumanvid}, comprising 1,000 hours of training data and evaluation sets of 200 short and 200 long videos.
Across diverse scenarios, \modelname improves identity similarity while preserving target attributes and maintaining high video quality.
It processes videos at 20 FPS on a single NVIDIA 4090 GPU and maintains stable generation for videos up to 10 minutes.





Our contributions are summarized as follows:
\begin{itemize}
    \item We identify and examine two task-specific properties of reference-guided video face swapping: latent flow-matching supervision is weakly sensitive to identity changes, while target-provided visual information allows generation with fewer spatial tokens and shorter temporal contexts.

    \item We introduce \modelname, which combines \lossname, \roiname, and \contextname to enable real-time streaming video face swapping. It achieves 20 FPS on a single NVIDIA 4090 GPU with stable generation on videos up to 10 minutes long.

    \item We construct \datasetname, comprising 1,000 hours of cinematic training data derived from OpenHumanVid and evaluation sets of 200 short and 200 long videos for assessing identity fidelity, target-attribute preservation, visual quality, and temporal stability.
\end{itemize}

\section{Preliminaries}

\subsection{Flow Matching}
\label{sec:flow_matching}


Flow matching (FM)~\citep{lipman2022flow} learns a continuous velocity field that transports samples from a Gaussian noise distribution to a target distribution defined in either data or latent space.
Let \(x_1\) and \(x_0\) denote a clean video latent and random Gaussian noise in the same latent space, respectively.
The intermediate latent along the linear probability path is defined as
\begin{equation}
    x_t=(1-\sigma_t)x_1+\sigma_t x_0,
    \label{eq:flow_path}
\end{equation}
where $\sigma_t = 1 - t$ denotes the noise level at a continuous timestep $t \in [0, 1]$.
Thus, \(t=0\) (\(\sigma_t=1\)) corresponds to pure noise and \(t=1\) (\(\sigma_t=0\)) corresponds to clean latent.
Given a trainable velocity model $v_\theta$, the flow matching objective is defined as
\begin{equation}
    \mathcal{L}_{\mathrm{flow}}
    =
    \mathbb{E}_{(x_1,c), x_0 \sim \mathcal{N}(0, I),t\sim \mathcal{U}(0,1)}
    \left[
        \left\|
            v_\theta(x_t,c,t)-v_t^\star
        \right\|_2^2
    \right],
    \label{eq:flow_loss}
\end{equation}
where $(x_1,c)$ is a training pair of a clean video latent and its conditioning signal, sampled from the training dataset.
$v_t^\star=\frac{\mathrm{d}x_t}{\mathrm{d}t}=x_1-x_0$ is the target velocity.
$\mathcal{N}(0,I)$ and $\mathcal{U}(0,1)$ denote the standard Gaussian distribution and the uniform distribution over $[0,1]$, respectively.

\subsection{Reference-guided Video Face Swapping}
\label{sec:reference_guide}

Reference-guided video face swapping conditions generation on the target video to preserve its expression, pose, illumination, motion, and background.

\textbf{Model conditioning.}
Reference-guided models incorporate source-image and target-video representations as complementary conditions in the denoising network.
A representative design is DreamID-V~\citep{guo2026dreamid}, which builds on a Wan Transformer~\citep{wan2025wan} and injects video and source-image conditions through channel-wise and token-wise concatenation, respectively.
Formally, at each time step $t$, its velocity predictor $v_\theta(x_t,c,t)$ takes the noisy latent $x_t$  together with conditioning information $c=\mathcal{C}(I, V)$ derived from the source image $I$ and target video $V$.
To construct these conditioning inputs, a frozen VAE encoder $\mathcal{E}_{\phi}$ first encodes the target video and its face mask $M$ into $x_v$ and $x_m$, respectively.
Then, the noisy latent $x_t$ is concatenated with these clean video conditions and embedded into video tokens:
\begin{equation}
H_{\mathrm{vid}}
=
\mathcal{P}_{\mathrm{3D}}\!\left(
\operatorname{Concat}_{\mathrm{ch}}[x_t,x_v,x_m]
\right)
\in \mathbb{R}^{(hwF)\times d},
\label{eq:method_video_tokens}
\end{equation}
where $\mathcal{P}_{\mathrm{3D}}$ denotes the 3D patch embedding layer, $h$, $w$, and $F$ are the height, width, and temporal length of the resulting token grid, and $d$ is the dimension of token feature. $\operatorname{Concat}_{\mathrm{ch}}$ denotes channel-wise concatenation.
The grid is flattened into a sequence, with each token retaining its corresponding spatial and temporal position.
The source image is encoded by the same VAE and a separate 2D patch embedding $\mathcal{P}_{\mathrm{2D}}$, yielding identity tokens that are concatenated with the video tokens:
\begin{equation}
H_{\mathrm{id}}
=
\mathcal{P}_{\mathrm{2D}}\!\left(\mathcal{E}_{\phi}(I)\right),
\qquad
H
=
\operatorname{Concat}_{\mathrm{seq}}
[H_{\mathrm{id}},H_{\mathrm{vid}}],
\label{eq:method_joint_tokens}
\end{equation}
where $\operatorname{Concat}_{\mathrm{seq}}$ denotes sequence-wise concatenation.
The Transformer then processes $H$ through bidirectional self-attention to predict the final output of the velocity predictor $v_{\theta}(x_t,c,t)$.

\textbf{Paired data and reversing supervision.}
Since paired face-swapping videos are difficult to obtain, LivingSwap~\citep{luo2026preserving} and DreamID-V~\citep{guo2026dreamid} construct paired data using pretrained face-swapping pipelines.
Following the notation of DreamID-V, let $(I_r,V_r)$ denote a real image-video pair with identity $r$, and $I_g$ depict identity $g$.
A face-swapping pipeline $\mathcal{S}$ produces
\begin{equation}
V_g=\mathcal{S}(I_g,V_r),
\label{eq:synthetic_pair}
\end{equation}
where $V_g$ carries identity $g$ while approximately preserving the non-identity attributes and temporal dynamics of $V_r$.
The resulting quadruplet $(I_r,V_r,I_g,V_g)$ supports role-reversed training: $I_r$ provides the source identity, $V_g$ serves as the target reference video, and $V_r$ provides ground-truth supervision.
This reversal avoids using synthesis artifacts as ground-truth targets.
Training follows Eq.~\ref{eq:flow_loss} with $x_1=\mathcal{E}_{\phi}(V_r)$ and $c=c_{\mathrm{swap}}=\mathcal{C}(I_r,V_g)$, with $x_t$ constructed from $x_1$ using Eq.~\ref{eq:flow_path}.

\section{Task-Specific Analysis}
\label{sec:task_analysis}

The difficulty of combining faithful identity transfer, high visual quality, and real-time long-video generation motivates us to revisit video face swapping through its task-specific structure.
Compared with general-purpose video editing, reference-guided face swapping has two distinctive properties.

First, face swapping has a restricted editing scope: only facial identity should change, while all other facial attributes and content outside the face region should remain consistent with the target video.
However, the flow matching objective optimizes a reconstruction loss over the latent representation of the entire image, which may be insufficiently sensitive to identity changes when most visual content remains unchanged.
To quantify this potential mismatch, we investigate 
\textit{how strongly the flow-matching objective penalizes identity-transfer failures} \textbf{(Q1)}.



Second, the desired output is densely specified by visual inputs: the source image provides the identity, while the target video supplies not only non-identity facial attributes within the edited region but also the surrounding content and temporal dynamics.
Nevertheless, existing reference-guided methods generate full frames, including regions that require no editing, and attend over extended temporal contexts to maintain consistency.
To examine this potential computational redundancy, we investigate \textit{how much spatial and temporal context is actually required to preserve fidelity} \textbf{(Q2)}.

We investigate these questions through two controlled analysis experiments.

\begin{figure*}[t]
    \centering
    \includegraphics[width=.95\linewidth]{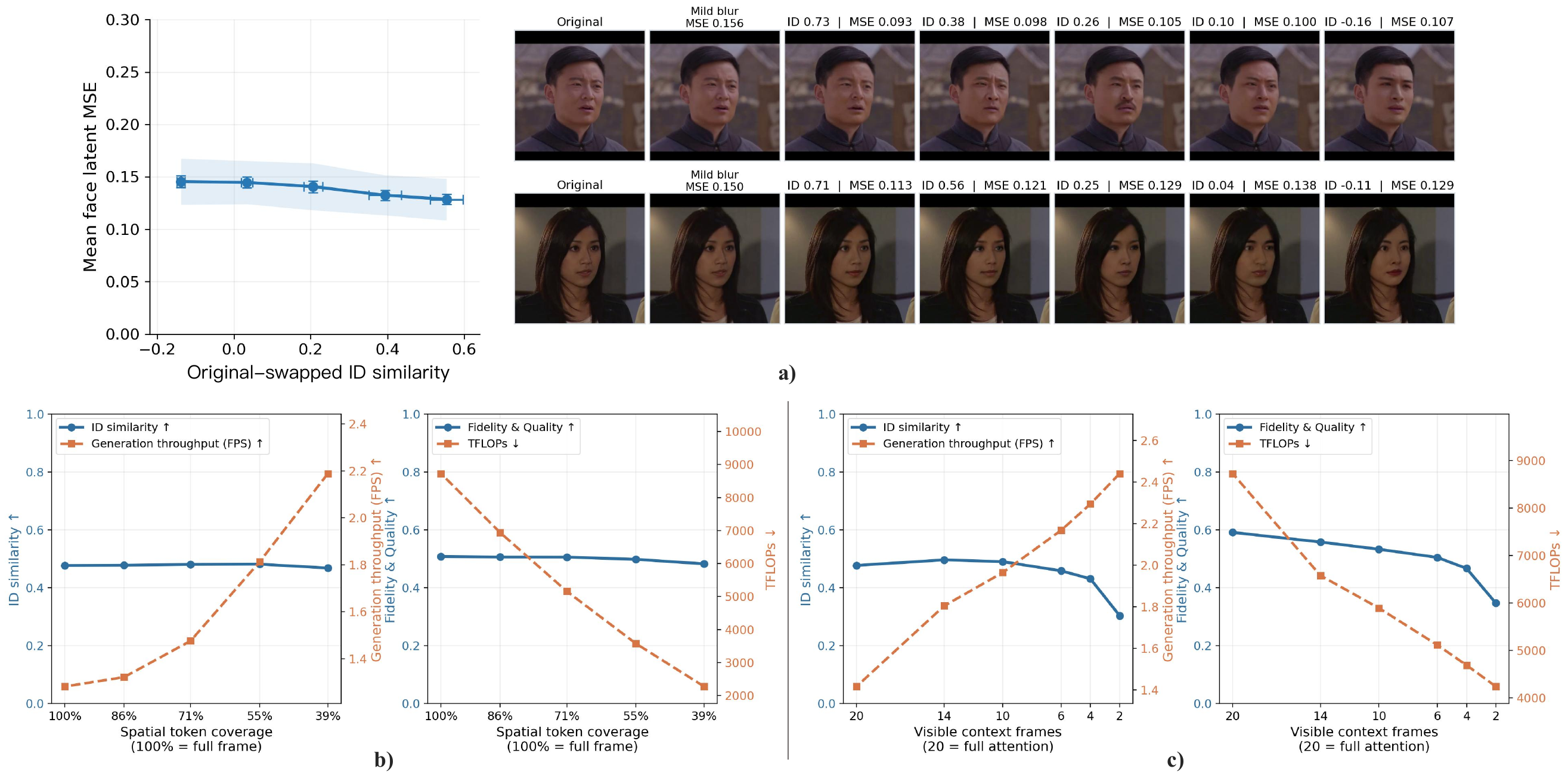}
    \vspace{-10pt}
    \caption{
        Task-specific analysis of reference-guided video face swapping.
        (a) Latent flow-matching supervision is weakly sensitive to identity changes.
        (b) Face-region processing and (c) short temporal context preserve most generation quality while substantially reducing computation.
    }
    \label{fig:insight}
    \vspace{-10pt}
\end{figure*}

\subsection{Identity Sensitivity of Flow-Matching Supervision}
\label{sec:id_insight}

\textbf{Setup.}
Under the linear probability path in Eq.~\ref{eq:flow_path}, the predicted velocity determines the clean-latent estimate
\begin{equation}
    \hat{x}_1
    =
    x_t+\sigma_t v_\theta(x_t,c_{\mathrm{swap}},t).
    \label{eq:predicted_clean_latent}
\end{equation}
Because \(x_1=x_t+\sigma_t v_t^\star\), for \(\sigma_t>0\),
\begin{equation}
    \left\|
        v_\theta(x_t,c_{\mathrm{swap}},t)-v_t^\star
    \right\|_2^2
    =
    \frac{1}{\sigma_t^2}
    \left\|
        \hat{x}_1-x_1
    \right\|_2^2.
    \label{eq:flow_as_latent_reconstruction}
\end{equation}
Thus, at fixed noise, flow matching is equivalent to a scaled clean-latent reconstruction objective.
Under role-reversed training, a model that ignores $I_r$ and reconstructs the target video $V_g$ is penalized according to the latent distance between $V_g$ and the ground-truth video $V_r$.
This latent distance therefore provides a diagnostic of the flow-matching supervision sensitivity to identity differences.
We randomly sample 64 real videos $V_r$.
For each video, we select five identity images $\{I_g^{(k)}\}_{k=1}^{5}$ with varying identity similarities to $I_r$, including $I_r$ as the self-swap control.
Following the paired-data construction used for training, we use InSwapper~\citep{facefusion2025} to generate \(V_g^{(k)}=\mathcal{S}(I_g^{(k)},V_r)\).
We encode $V_r$ and each $V_g^{(k)}$ using $\mathcal{E}_{\phi}$ to compute their face-region latent MSE following Eq.~\ref{eq:flow_as_latent_reconstruction}, and measure identity similarity between the two videos.
To contextualize the MSE scale, the qualitative visualization additionally includes a mildly blurred identity-preserving version of \(V_r\) as a reference.

\textbf{Results.}
As shown in Fig.~\ref{fig:insight}(a), identity similarity changes by more than \(0.75\) across the constructed pairs, whereas mean face-region latent MSE varies by less than \(0.05\).
The identity-preserving self-swap (third column) and mild-blur controls (second column) contextualize this error scale: visually minor deviations already produce latent MSEs around \(0.15\), comparable to or larger than those caused by visible identity changes (last column).
This mismatch is consistent with the observation that swapping identity leaves most facial content, including expression, pose, and illumination, largely unchanged.
At a fixed noise level, the corresponding flow-matching penalty therefore changes little as the identity discrepancy increases.
These results suggest that flow-matching supervision can support identity transfer but provides only a weak signal for identity changes.

\subsection{Spatial and Temporal Context Requirements}
\label{sec:context_insight}




\textbf{Setup.}
We examine the spatial and temporal context required by DreamID-V~\citep{guo2026dreamid} using the same 64 evaluation cases.
We independently vary the spatial coverage and temporal attention range of the video tokens $H_{\mathrm{vid}}\in\mathbb{R}^{(hwF)\times d}$ defined in Eq.~\ref{eq:method_video_tokens}.

Spatially, we restrict token selection along the $h\times w$ spatial grid while retaining all $F$ temporal frames.
We progressively narrow the retained spatial region from the full frame to an enlarged face editing region.
For the retained video-token indices $\mathcal{J}_{\rho}$, the selected token sequence and its spatial coverage are given by
\begin{equation}
H_{\mathrm{vid}}^{(\rho)}
=
H_{\mathrm{vid}}[\mathcal{J}_{\rho}],
\qquad
\rho
=
\frac{|\mathcal{J}_{\rho}|}{hwF},
\label{eq:spatial_context_analysis}
\end{equation}
where $\rho = 100\%$ corresponds to full-frame processing.
Temporally, we use an attention mask to restrict each video token's temporal context to $K$ neighboring latent frames.
We vary $K$ over $\{20,14,10,6,4,2\}$, with $K=20=F-1$ corresponding to full temporal bidirectional attention.

We report identity similarity, expression and lighting errors, imaging quality, generation throughput in FPS, and computational cost in TFLOPs.
For visualization, Fig.~\ref{fig:insight} aggregates expression error, lighting error, and imaging quality into a normalized fidelity-and-quality score.

\textbf{Results.}
Spatially, Fig.~\ref{fig:insight}(b) shows that reducing the retained-token ratio from \(100\%\) to \(39\%\) largely preserves identity similarity and the fidelity-and-quality score, while increasing generation throughput from approximately \(1.3\) to \(2.2\) FPS and reducing computation from \(8{,}900\) to \(2{,}200\) TFLOPs.
This result indicates that background tokens contribute substantial computation but limited information once the target face region is available in reference-guided methods.

Temporally, Fig.~\ref{fig:insight}(c) shows that reducing the visible context from full attention to six neighboring latent frames causes only moderate quality degradation while increasing generation throughput and reducing computational cost.
Performance drops sharply with fewer neighboring frames, particularly at $K=2$.
These results suggest that a short temporal context is sufficient for reference-guided generation, as the target video already provides the content and motion cues needed for long-range consistency.
The temporal context therefore serves primarily to maintain local continuity.

\textbf{Summary.}
Together, the two experiments motivate the following design principles:
\begin{itemize}
    \item \textbf{Explicit identity supervision.}
    Training should directly supervise source identity to complement the weak sensitivity of latent reconstruction to identity changes.

    \item \textbf{Focused spatial and temporal computation.}
    Generation should exploit the target-provided content and motion to concentrate computation on face-region tokens and a short temporal context.
\end{itemize}

\section{Method}
\label{sec:method}

\begin{figure*}[t]
    \centering
    \includegraphics[width=.95\linewidth]{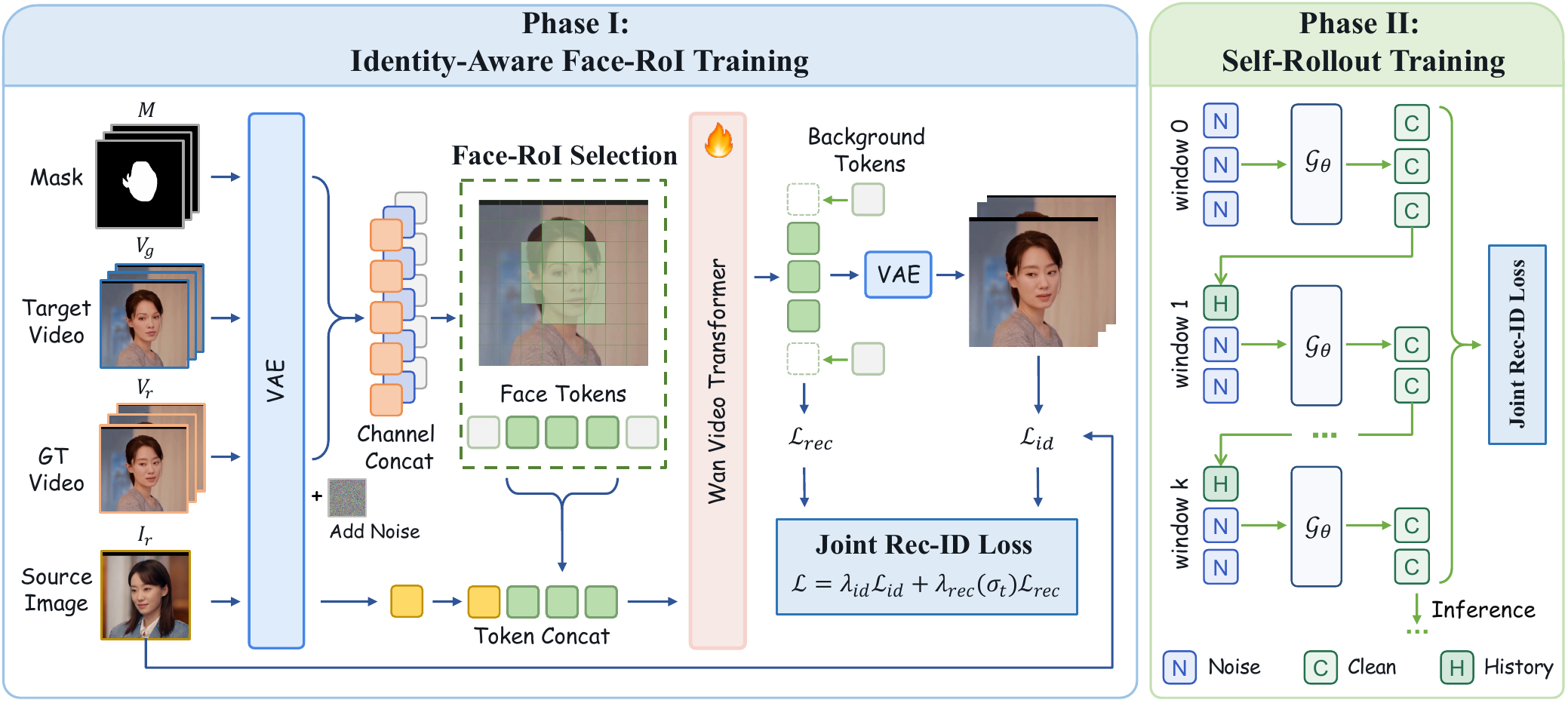}
    \vspace{-10pt}
    \caption{
    Overview of \modelname.
    Phase I restricts video processing to face-RoI tokens and balances latent reconstruction with identity supervision through noise-dependent weighting.
    Phase II fine-tunes the model through self-rollout with short generated histories, enabling streaming inference while retaining bidirectional generation within each chunk.
    }
    \vspace{-10pt}
    \label{fig:method}
\end{figure*}

\subsection{Overview}
\label{sec:method_overview}

Our pipeline comprises data preparation, two training phases, and streaming inference.
We first construct the reversed pairs as in App.~\ref{sec:data_preparation}.
Phase I adapts a reference-guided video flow model with \lossname and \roiname.
Phase II employs \contextname, exposing the model to its own generated short history for autoregressive long-video inference (Fig.~\ref{fig:method}).

\subsection{Phase I: Identity-Aware Face-RoI Training}
\label{sec:phase_one}

We adopt the architecture and reference-guided training setup described in Sec.~\ref{sec:reference_guide}, initializing the model with pretrained DreamID-V weights~\citep{guo2026dreamid}.
Guided by our task-specific analysis, we introduce two modifications.

\textbf{\roiname.}
The analysis in Sec.~\ref{sec:context_insight} suggests that processing background tokens incurs substantial computational cost while providing limited performance gains.
We therefore restrict spatial token coverage to an enlarged face region.
Specifically, we downsample the binary face masks $M$ to the latent grid using max pooling and spatially dilate the resulting masks to define the compute region.
We select all patches intersecting this region, denoting their token indices by $\mathcal{J}_{\mathrm{roi}}$ and the latent positions they cover by $\Omega_{\mathrm{roi}}$.
As illustrated in Fig.~\ref{fig:method} (left), the Transformer input is
\begin{equation}
H_{\mathrm{roi}}
=
\operatorname{Concat}_{\mathrm{seq}}
\left[
H_{\mathrm{id}},
H_{\mathrm{vid}}[\mathcal{J}_{\mathrm{roi}}]
\right].
\label{eq:method_roi_tokens}
\end{equation}
Selected video tokens retain their original spatiotemporal positional encodings.
To support batched training, we pad sequences to the maximum length within each batch and mask out padding tokens in attention.
We recover the clean RoI latent estimate $\hat{x}_{1,\mathrm{roi}}$ from the predicted velocity following Eq.~\ref{eq:predicted_clean_latent}.
This estimate is scattered back to the full latent grid and blended with the target-video latent $x_g=\mathcal{E}_{\phi}(V_g)$:
\begin{equation}
\tilde{x}_1
=
M_{\mathrm{blend}}\odot
\operatorname{Scatter}_{\Omega_{\mathrm{roi}}}
\left(\hat{x}_{1,\mathrm{roi}}\right)
+
(1-M_{\mathrm{blend}})\odot x_g,
\label{eq:method_roi_composition}
\end{equation}
where $\operatorname{Scatter}_{\Omega_{\mathrm{roi}}}$ restores the predicted latents to their original positions.
The feathered blending mask $M_{\mathrm{blend}}$ is derived from the same downsampled face masks, with weights transitioning smoothly from 1 in the inner face region to 0 outside the compute region.

\textbf{\lossname.}
As shown in Sec.~\ref{sec:id_insight}, latent reconstruction is relatively insensitive to facial identity changes.
We therefore combine clean-latent reconstruction with explicit identity supervision (Fig.~\ref{fig:method}, left).
To reduce training memory usage, we decode a small subset of frames from $\tilde{x}_1$ and evaluate their identity using a frozen face-recognition encoder $f_{\mathrm{id}}$~\citep{deng2019arcface}:
\begin{equation}
\mathcal{L}_{\mathrm{id}}
=
\frac{1}{|\mathcal{K}|}
\sum_{k\in\mathcal{K}}
\left[
1-
\cos\!\left(
f_{\mathrm{id}}(\hat{I}_k),
f_{\mathrm{id}}(I_r)
\right)
\right],
\label{eq:method_id_loss}
\end{equation}
where $\hat{I}_k$ is a decoded frame and $\mathcal{K}$ indexes the selected frames with valid face detections.
Before blending, the predicted clean RoI latent is supervised against the corresponding ground-truth latent:
\begin{equation}
\mathcal{L}_{\mathrm{rec}}
=
\left\|
\hat{x}_{1,\mathrm{roi}}-x_{1,\mathrm{roi}}
\right\|_2^2,
\label{eq:method_latent_loss}
\end{equation}
where $x_{1,\mathrm{roi}}$ is the restriction of $x_1$ to $\Omega_{\mathrm{roi}}$.
Phase I minimizes
\begin{equation}
\mathcal{L}
=
\mathbb{E}_{(I_r,V_g,V_r),x_0,t}
\left[
\lambda_{\mathrm{id}}\mathcal{L}_{\mathrm{id}}
+
\lambda_{\mathrm{rec}}(\sigma_t)\mathcal{L}_{\mathrm{rec}}
\right],
\label{eq:method_phase_one_objective}
\end{equation}
where the expectation is over role-reversed training samples, Gaussian noise, and flow time.
The identity weight $\lambda_{\mathrm{id}}$ remains fixed, while the reconstruction weight depends on the noise level:
\begin{equation}
\lambda_{\mathrm{rec}}(\sigma_t)
=
\begin{cases}
\lambda_{\mathrm{high}}, & \sigma_t>\tau,\\
\lambda_{\mathrm{low}},  & \sigma_t\leq\tau,
\end{cases}
\qquad
\lambda_{\mathrm{high}}>\lambda_{\mathrm{low}}.
\label{eq:method_latent_weight}
\end{equation}
This schedule emphasizes reconstruction supervision at high noise levels to guide clean-latent recovery, including facial identity, while giving identity supervision greater relative weight at low noise levels to refine identity-specific facial details (App.~\ref{sec:ab_loss_weight}).

\subsection{Phase II: Short-Context Rollout Training}
\label{sec:phase_two}

Recent forcing-based methods adapt bidirectional video models for streaming generation through causal autoregressive modeling~\citep{yin2025slow,huang2025selfforcing,zhu2026causal}.
Our analysis in Sec.~\ref{sec:context_insight} motivates a simpler adaptation for reference-guided face swapping: the target video supplies content and motion cues, so a short generated history can provide context for local temporal continuity.
We therefore retain bidirectional attention within each window and condition the generation of new frames on this short history.

Let $N_{\mathrm{ctx}}$ and $N_{\mathrm{gen}}$ denote the numbers of history and newly generated latent frames, respectively.
For generation window $n$, we take the latest $N_{\mathrm{ctx}}$ latent frames of previous output $\tilde{x}_1^{(n-1)}$ as $x_{\mathrm{ctx}}^{(n)}$ and generate the next chunk as
\begin{equation}
\hat{x}_{1,\mathrm{roi}}^{(n)}
=
\mathcal{G}_{\theta}\!\left(
x_0^{(n)};
x_{\mathrm{ctx}}^{(n)},
c_{\mathrm{swap}}^{(n)}
\right),
\label{eq:method_short_context_generation}
\end{equation}
where $x_0^{(n)}$ contains Gaussian noise for $N_{\mathrm{gen}}$ latent frames and $\mathcal{G}_{\theta}$ denotes the multi-step denoising process using the velocity model $v_{\theta}$.
The condition $c_{\mathrm{swap}}^{(n)}$ includes the source image and aligned target-video and mask inputs covering both the history and new frames.
History latents remain fixed at $\sigma=0$ throughout denoising.
The predicted RoI latents are composited using Eq.~\ref{eq:method_roi_composition}, and the latest composited frames provide history for the next window.
The first window generates $N_{\mathrm{ctx}}+N_{\mathrm{gen}}$ latent frames without history.
We use $N_{\mathrm{ctx}}=3$ and $N_{\mathrm{gen}}=9$ to balance generation quality, throughput, and latency (App.~\ref{sec:ab_window}).

Training with ground-truth history creates a mismatch with streaming inference, where prediction errors can propagate through the generated history and accumulate across chunks.
Inspired by Self Forcing~\citep{huang2025selfforcing}, we fine-tune the Phase-I model through autoregressive self-rollout, exposing it to its own generated history during training (Fig.~\ref{fig:method}, right).
We use a 4-step denoising schedule, as the Phase-I model achieves comparable identity and target-attribute performance with 4 and 25 steps at substantially higher throughput (App.~\ref{sec:ab_phase}).
Following the stochastic gradient-truncation strategy of Self Forcing, we randomly select a denoising step as the stopping point and backpropagate only through the clean-latent prediction at that step.
Gradients are stopped through both the generated history and preceding denoising steps.
Rather than a distribution-matching objective, we use the Phase-I identity and Face-RoI reconstruction losses in Eq.~\ref{eq:method_phase_one_objective} to supervise the newly generated frames, retaining the noise-dependent weights in Eq.~\ref{eq:method_latent_weight}.

\section{Experiments}

\begin{table*}[t]
\centering
\caption{
Quantitative comparison on short and long videos.
\modelname combines effective identity transfer with strong target-attribute preservation and video quality at over 20 FPS.
}
\label{tab:quantitative_comparison}
\begingroup
\setlength{\tabcolsep}{3pt}
\renewcommand{\arraystretch}{1.08}
\resizebox{0.8\textwidth}{!}{%
\begin{tabular}{l*{8}{c}}
\toprule
\multirow{2}{*}{Method}
& \multicolumn{1}{c}{Identity}
& \multicolumn{4}{c}{Attribute Preservation}
& \multicolumn{1}{c}{Quality}
& \multicolumn{2}{c}{Efficiency} \\
\cmidrule(lr){2-2}
\cmidrule(lr){3-6}
\cmidrule(lr){7-7}
\cmidrule(lr){8-9}
& ID Sim.$\uparrow$
& Pose$\downarrow$
& Expr.$\downarrow$
& Light$\downarrow$
& Gaze$\uparrow$
& FVD$\downarrow$
& FPS$\uparrow$
& Latency (s)$\downarrow$ \\

\midrule
\multicolumn{9}{c}{\textbf{Short videos}} \\

BlendFace
& 0.533
& 6.218 & \textbf{2.471} & 0.301 & \underline{0.875}
& \underline{130.77}
& 16.548 & \underline{0.276} \\

InSwapper
& \textbf{0.735}
& \underline{6.084} & 2.624 & \underline{0.252} & 0.873
& 138.09
& \textbf{28.621} & \textbf{0.050} \\

FaceAdapter
& 0.129
& 8.563 & 2.681 & 0.329 & 0.823
& 257.99
& 0.353 & 2.833 \\

VividFace
& 0.175
& 8.870 & 2.695 & 0.314 & 0.786
& 204.15
& 0.283 & 28.267 \\

VFace
& 0.373
& 11.258 & 3.004 & 0.366 & 0.761
& 252.24
& 0.129 & 20.662 \\

DreamID-V
& \underline{0.734}
& 9.338 & 3.009 & 0.509 & 0.764
& 231.00
& 1.815 & 44.891 \\

\cmidrule(lr){1-9}
\modelname (Ours)
& 0.577
& \textbf{5.354} & \underline{2.484}
& \textbf{0.242} & \textbf{0.881}
& \textbf{129.09}
& \underline{21.174} & 1.782 \\

\midrule
\multicolumn{9}{c}{\textbf{Long videos}} \\

BlendFace
& 0.533
& 7.173 & \textbf{2.464} & 0.306 & \underline{0.867}
& 154.05
& 16.548 & \underline{0.276} \\

InSwapper
& \textbf{0.723}
& \underline{6.701} & 2.584 & \underline{0.248} & 0.867
& \underline{147.01}
& \textbf{28.621} & \textbf{0.050} \\

DreamID-V
& \underline{0.715}
& 10.827 & 2.967 & 0.489 & 0.798
& 249.41
& 1.815 & 44.891 \\

\cmidrule(lr){1-9}
\modelname (Ours)
& 0.576
& \textbf{6.336} & \underline{2.466}
& \textbf{0.238} & \textbf{0.881}
& \textbf{136.23}
& \underline{21.174} & 1.782 \\

\bottomrule
\end{tabular}%
}
\endgroup
\end{table*}

\begin{figure*}[t]
    \centering
    \vspace{-10pt}
    \includegraphics[width=.85\linewidth]{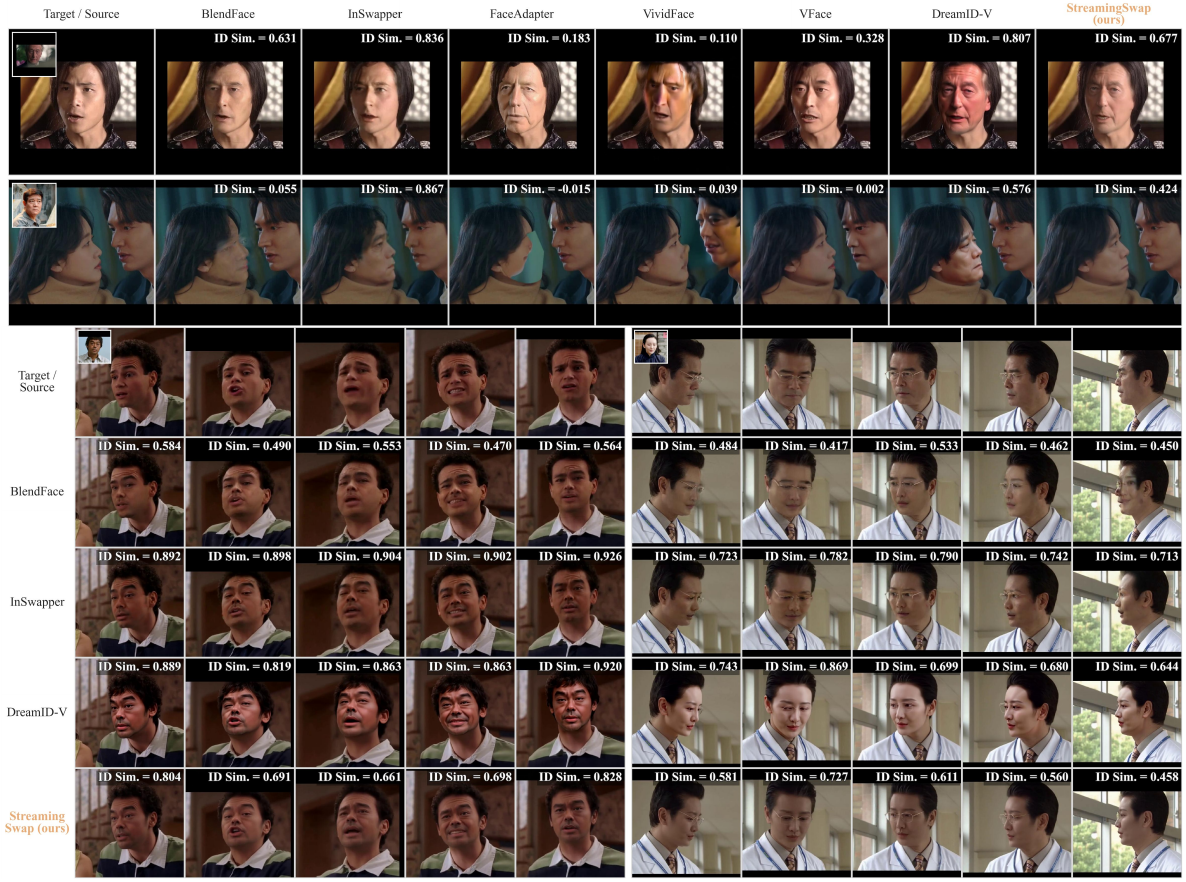}
    \vspace{-10pt}
    \caption{
    Qualitative comparison on short and long videos.
    Higher identity similarity can coexist with reduced target fidelity and realism, whereas \modelname balances them.
}
    \label{fig:compare1}
    \vspace{-10pt}
\end{figure*}



We initialize \modelname from DreamID-V~\citep{guo2026dreamid} and train each phase for 1,000 iterations.
We evaluate identity similarity, target-attribute preservation, FVD, and end-to-end inference efficiency.
All speed measurements use a single NVIDIA RTX 4090 GPU.
Detailed metrics and implementation settings are provided in App.~\ref{sec:exp_settings}.

\subsection{Comparisons with Existing Methods}
\label{sec:exp_compare}

We compare \modelname with two representative GAN-based methods and four diffusion-based methods that can run on a single NVIDIA RTX 4090 GPU, with quantitative and qualitative results in Table~\ref{tab:quantitative_comparison} and Fig.~\ref{fig:compare1}.
Long-video comparisons include three strong-performing methods.
Across both benchmarks, \modelname achieves effective identity transfer while preserving target fidelity and visual realism.
Although InSwapper and DreamID-V achieve higher identity similarity, the frame-level scores in Fig.~\ref{fig:compare1} illustrate that higher embedding similarity does not necessarily imply better perceptual quality or target fidelity.
DreamID-V exhibits pronounced changes in lighting and head pose despite high ID Sim. scores, while InSwapper's high scores coexist with less realism in the displayed examples.
In contrast, \modelname maintains recognizable source identity while more closely following the target lighting, expressions, and head poses.
These comparisons highlight the importance of balancing identity similarity and target-attribute reconstruction, as further examined in App.~\ref{sec:ab_loss_weight}. 
Overall, \modelname offers a balanced quality--efficiency, exceeding 20 FPS and achieving substantially higher throughput than the evaluated diffusion baselines.

\subsection{Ablation of Identity-aware Supervision and Efficient Context}
\label{sec:ab_main}













\begin{table*}[t]
\centering
\caption{
Ablation of the main components.
Phase-I variants remove identity supervision or Face-RoI.
The Phase-II variant replaces rollout training with ground-truth-history conditioning.
}
\label{tab:component_ablation}

\begingroup
\setlength{\tabcolsep}{6pt}
\renewcommand{\arraystretch}{1.08}

\resizebox{0.8\textwidth}{!}{%
\begin{tabular}{lcccccc}
\toprule
Variant
& ID Sim.$\uparrow$
& Pose$\downarrow$
& Expr.$\downarrow$
& Light$\downarrow$
& Gaze$\uparrow$
& FPS$\uparrow$ \\
\midrule

\multicolumn{7}{c}{Phase I on short videos} \\
\midrule

DreamID-V (Initialization, no CFG)
& 0.445 & 8.504 & 2.695 & 0.427 & 0.798
& 1.815 \\

w/o ID supervision
& 0.241 & 4.813 & 2.014 & 0.212 & 0.883
& 6.702 \\

w/o Face-RoI
& 0.564 & 5.647 & 2.516 & 0.276 & 0.861
& 2.355 \\


\modelname (Phase I)
& 0.573 & 5.723 & 2.526 & 0.266 & 0.873
& 6.702 \\

\midrule
\multicolumn{7}{c}{Phase II on long videos} \\
\midrule

w/o self-rollout
& 0.590 & 6.751 & 2.524 & 0.259 & 0.875
& 21.174 \\

\modelname (Phase II)
& 0.576 & 6.336 & 2.466 & 0.238 & 0.881
& 21.174 \\

\bottomrule
\end{tabular}%
}
\endgroup
\end{table*}

\begin{figure*}[t]
    \centering
    \vspace{-10pt}
    \includegraphics[width=.95\linewidth]{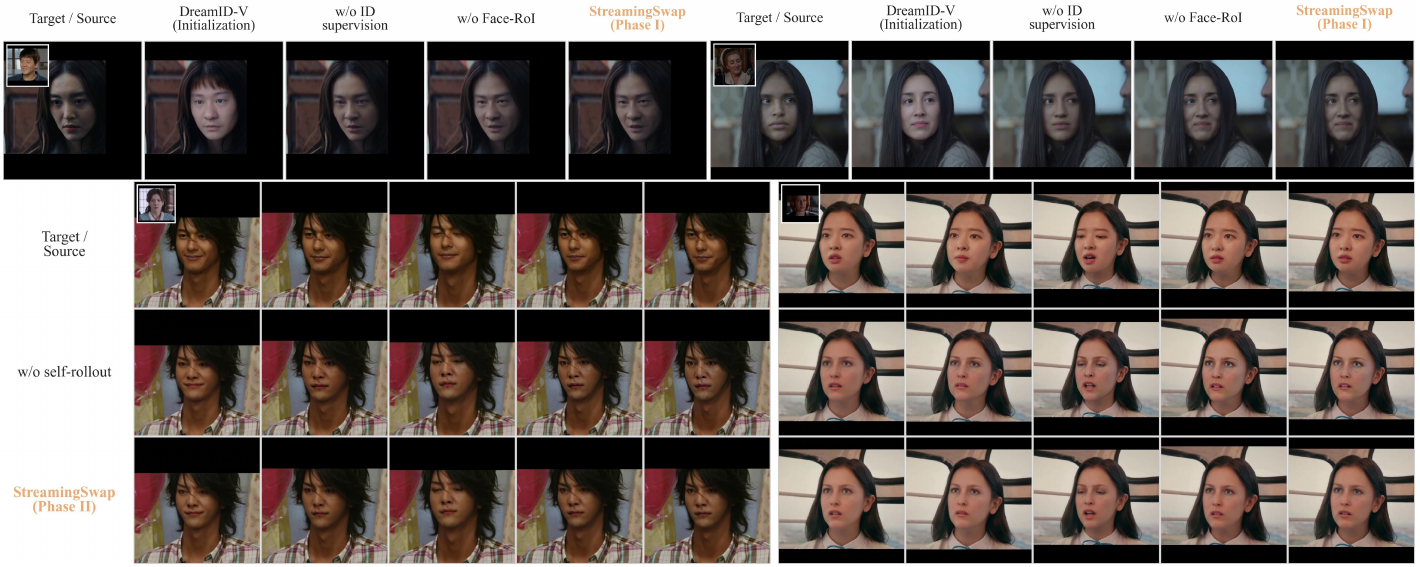}
    \vspace{-10pt}
    \caption{
        Qualitative ablation of \modelname.
        Top: identity supervision strengthens source-identity transfer, while Face-RoI preserves visual quality.
        Bottom: self-rollout aligns training with streaming inference, mitigating error accumulation and progressive visual degradation.
    }
    \vspace{-10pt}
    \label{fig:component_ablation}
\end{figure*}

\textbf{Setup.}
We separately remove identity supervision and Face-RoI in Phase I, evaluating all variants with 25 denoising steps.
For Phase II, we replace self-rollout with ground-truth-history conditioning and evaluate both variants using four-step inference.
Table~\ref{tab:component_ablation} and Fig.~\ref{fig:component_ablation} summarize the results.


\textbf{Identity supervision.}
Removing identity supervision weakens source-identity transfer despite improving target-attribute scores, as also reflected in the visual examples.
This confirms the analysis in Sec.~\ref{sec:id_insight}: flow matching supervision alone is insufficient to ensure the intended identity edit.

\textbf{Face-RoI processing.}
Face-RoI substantially increases throughput while maintaining comparable identity and target-attribute scores.
This is consistent with the analysis in Sec.~\ref{sec:context_insight}: background tokens add substantial computation but little useful information.

\textbf{Self-rollout training.}
The results in Fig.~\ref{fig:component_ablation} suggest that self-rollout training mitigates error accumulation across streaming windows by exposing the model to its own generated history.
\section{Conclusion}
\label{sec:conclusion}

We presented \modelname, a reference-guided framework for streaming video face swapping that aligns supervision and computation with the task requirements.
Our analysis motivates \lossname to strengthen identity transfer, alongside \roiname and \contextname to reduce spatial and temporal computation.
We also introduced \datasetname to support training and evaluation on short and long videos.
Experiments demonstrate competitive identity transfer and target fidelity at over 20 FPS on a single NVIDIA RTX 4090 GPU, with stable generation on videos up to 10 minutes long.
These results show that task-specific supervision and compact context can make reference-guided generation substantially more efficient.

\subsection*{AI Use Statement}

We used Codex to assist with Chinese--English translation, language refinement, interpretation of experimental results, and figure and table organization and LaTeX formatting.
We also used a pretrained face-swapping model to construct synthetic training videos, as described in Appendix~\ref{sec:data_preparation}.
The synthetic data were filtered using identity similarity and other quality metrics, followed by manual inspection of randomly sampled examples.
The authors take responsibility for the final content of this work, including all AI-assisted text.

\subsection*{Ethics Statement}

This work develops video face-swapping techniques for research and creative applications, including film production, AI-generated dramas, and live streaming.
We recognize that realistic identity manipulation can be misused for impersonation, misinformation, and non-consensual content creation, and that efficient streaming may increase these risks.
The additional application demos use video clips obtained from Bilibili and reference images located through Google Images.
The face-swapped outputs are synthetic research demonstrations and should not be interpreted as authentic recordings or endorsements by the depicted individuals.
We oppose deceptive and non-consensual uses of this technology and advocate obtaining appropriate permissions and consent, respecting privacy and content ownership, and clearly labeling manipulated media.

\subsection*{Reproducibility Statement}

The method section describes Face-RoI processing, joint reconstruction--identity supervision, short-context rollout training, and streaming inference.
Appendix~\ref{sec:data_preparation} details the data preparation pipeline.
Appendix~\ref{sec:exp_settings} provides additional training and evaluation settings, including model initialization, hyperparameters, sampling and window configurations, metric definitions, and hardware and acceleration settings.
We also specify the timing scope for end-to-end efficiency measurements.
Together, these descriptions support reproduction of the method and its experimental evaluation.

\bibliography{iclr2027_conference}
\bibliographystyle{iclr2027_conference}

\clearpage

\appendix


\section{Related Work}
\label{sec:relat}

Video face swapping transfers the identity of a source face to a target video while preserving target-specific attributes such as expression, pose, illumination, motion, and background.
Existing approaches can be broadly grouped into GAN-based methods, which emphasize efficient feed-forward synthesis, and diffusion-based methods, which leverage generative priors to improve visual quality and temporal coherence.

\subsection{GAN-based video face swapping}

GAN-based face swapping learns efficient feed-forward generators under adversarial supervision~\citep{goodfellow2014generative}.
Early systems used shared encoders with identity-specific decoders and therefore required separate training for each identity pair~\citep{deepfake}.
Later subject-agnostic methods inject source-identity features into target representations and improve identity transfer or identity-attribute disentanglement through specialized architectures, objectives, and identity encoders~\citep{li2019faceshifter,gao2021information,chen2020simswap,shiohara2023blendface}.
CanonSwap extends this direction to videos by modifying identity in a canonical space before restoring target motion~\citep{luo2025canonswap}.
InSwapper is a widely used one-shot model distributed through the InsightFace ecosystem, although its architecture and training procedure are not publicly documented~\cite{facefusion2025}.
While these methods offer efficient inference, most image-based variants process videos frame by frame and thus do not explicitly model temporal dependencies.

\subsection{Diffusion-based video face swapping}

The rapid development and application of diffusion models~\citep{ho2020denoising,rombach2022high,zhu2024unleashing,zhao20253d,zhao2026stroke3d} have enabled a second line of face-swapping methods with stronger generative priors.
One branch follows an inpainting-oriented formulation, in which the target face is masked or otherwise reconstructed from the source identity and explicit target-attribute conditions.
DiffSwap formulates face swapping as conditional inpainting and combines masked diffusion with identity and 3D facial guidance~\citep{zhao2023diffswap}.
Face-Adapter introduces lightweight modules for injecting source identity, spatial structure, and detailed target attributes into pretrained image diffusion models~\citep{han2024face}.
Extending this formulation to videos, DynamicFace decomposes facial conditions using 3D priors and incorporates temporal modeling, while HiFiVFS builds on a pretrained video diffusion model with disentangled target-attribute extraction and detailed identity injection~\citep{wang2025dynamicface,chen2024hifivfs}.
VividFace jointly trains on images and videos and introduces a video-oriented VAE, identity-attribute disentanglement data, and 3D facial guidance to improve temporal consistency and robustness~\citep{shao2024vividface}.
Related identity-control frameworks instead introduce lightweight conditional branches into pretrained video generators and can be adapted to face swapping, although face swapping is not their primary formulation~\cite{xue2026stand}.
These approaches improve synthesis quality and temporal modeling, but reconstructing a removed target face from auxiliary conditions can discard fine-grained appearance cues that are difficult to recover explicitly.

A more recent reference-guided branch retains the complete target video as a visual condition and learns face swapping from synthetically constructed pairs.
LivingSwap reverses GAN-generated training pairs so that real videos provide reconstruction targets, and uses edited keyframes together with temporal stitching to maintain identity across video chunks~\citep{luo2026preserving}.
DreamID-V constructs bidirectional identity supervision through SyncID-Pipe and trains a Diffusion Transformer with modality-aware conditioning, curriculum learning, and identity-coherence optimization~\citep{guo2026dreamid}.
By directly observing the target face, reference-guided models can preserve fine-grained expressions, illumination, makeup, and occlusions more faithfully.
However, access to the complete target face also allows the model to reconstruct most of the video while responding weakly to the source identity, and existing systems retain the computational cost of full-frame, chunk-based diffusion generation.
Our work follows the reference-guided formulation but studies a different boundary: identity-sensitive supervision and efficient streaming inference using face-region generation and a fixed-size generated history.

\section{Data Preparation}
\label{sec:data_preparation}

\begin{figure*}[t]
    \centering
    \includegraphics[width=\linewidth]{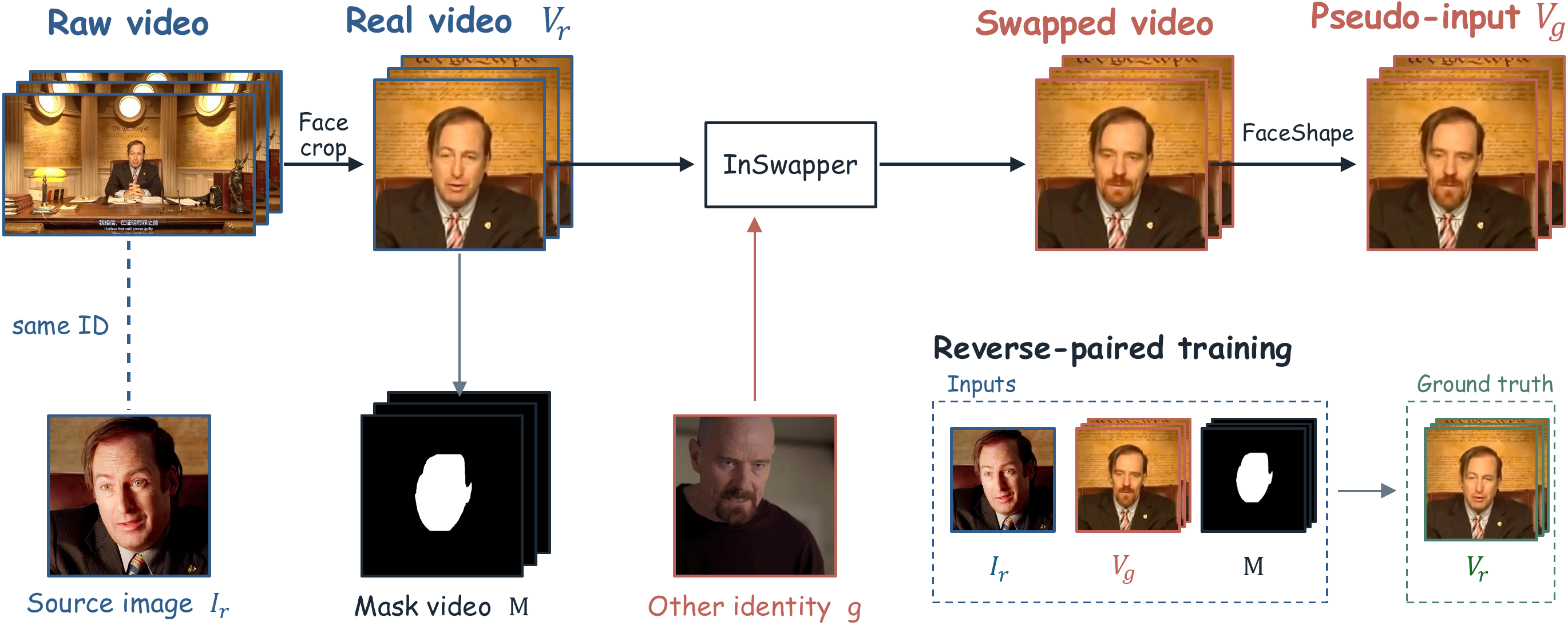}
    \vspace{-20pt}
    \caption{
    Data preparation and role-reversed training for \datasetname.
    We extract face-centered clips $V_r$ and aligned masks $M$, and select a source image $I_r$ with the same identity.
    InSwapper transfers another identity to $V_r$, followed by facial-contour warping to construct the pseudo-input $V_g$.
    Training conditions on $(I_r,V_g,M)$ to reconstruct the original real video $V_r$, avoiding synthesis artifacts as ground-truth targets.
}
    \label{fig:loss_schedule}
\end{figure*}

We construct \datasetname from OpenHumanVid~\citep{li2025openhumanvid}.
We detect and track faces, discard short or unreliable tracks, and obtain frame-aligned binary face masks by combining BiSeNet face parsing~\citep{yu2018bisenet} with InsightFace landmark convex hulls~\citep{insightface106}.
Source images are selected from cross-video identity clusters, yielding real image--video pairs \((I_r,V_r)\) of the same identity.
For each real video, a face-swapping model InSwapper~\citep{facefusion2025} transfers a different identity to produce the generated video \(V_g\) in Eq.~\ref{eq:synthetic_pair}.
We additionally apply temporally smoothed random facial-contour warps to \(V_g\), guided by MediaPipe landmarks~\citep{mediapipeFaceLandmarker}, to reduce residual geometric identity cues.
The resulting data are used in the reversed formulation, with the original real video retained as ground-truth supervision.

\section{Applications in Different Scenarios}
\label{sec:apply}

\begin{figure}[t]
    \centering
    \includegraphics[
        page=1,
        width=\textwidth,
        trim=45 20 45 20,
        clip
    ]{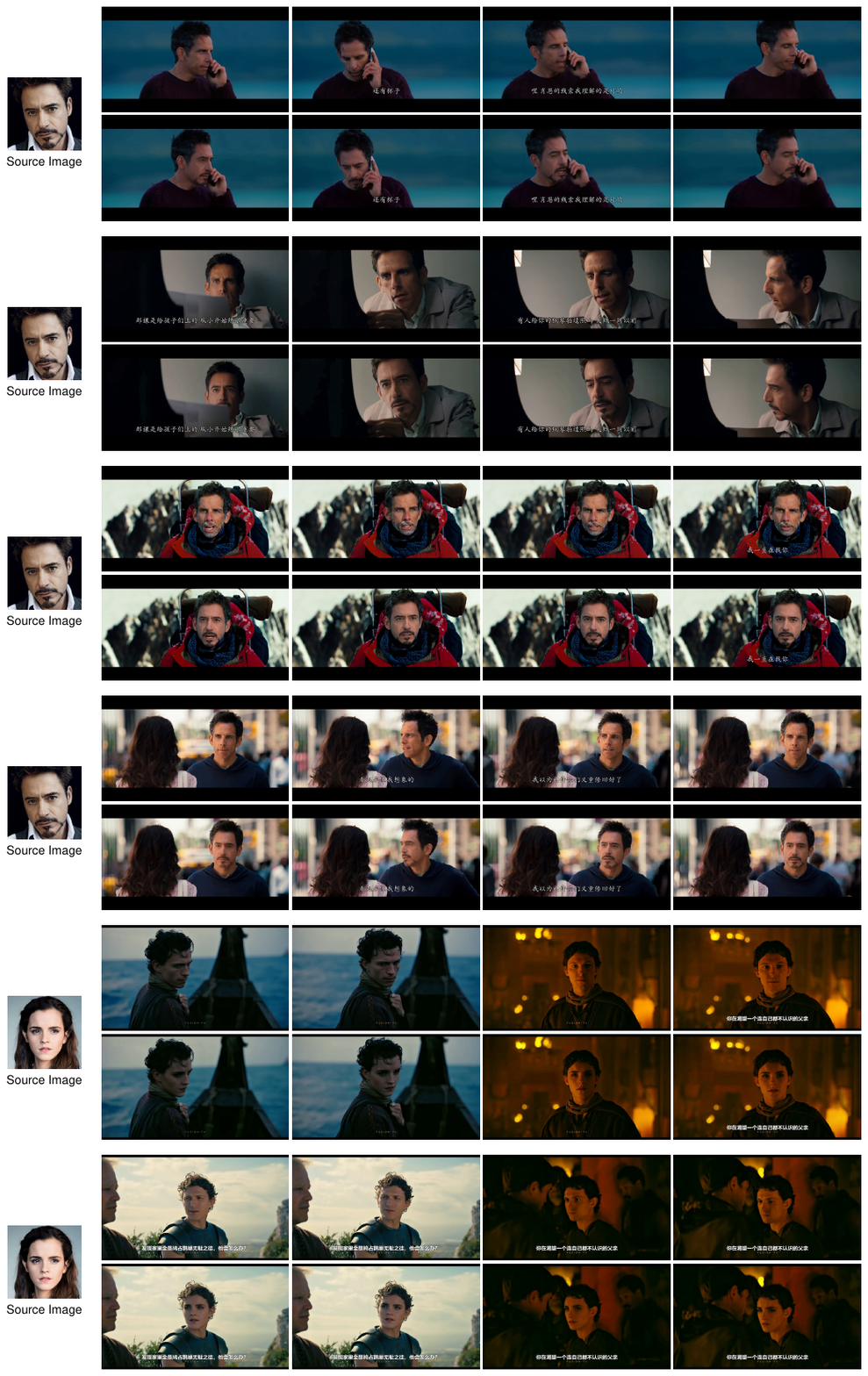}
    \caption{More qualitative results for movies.}
    \label{fig:app_movie}
\end{figure}

\begin{figure}[t]
    \centering
    \includegraphics[
        page=2,
        width=\textwidth,
        trim=25 125 25 125,
        clip
    ]{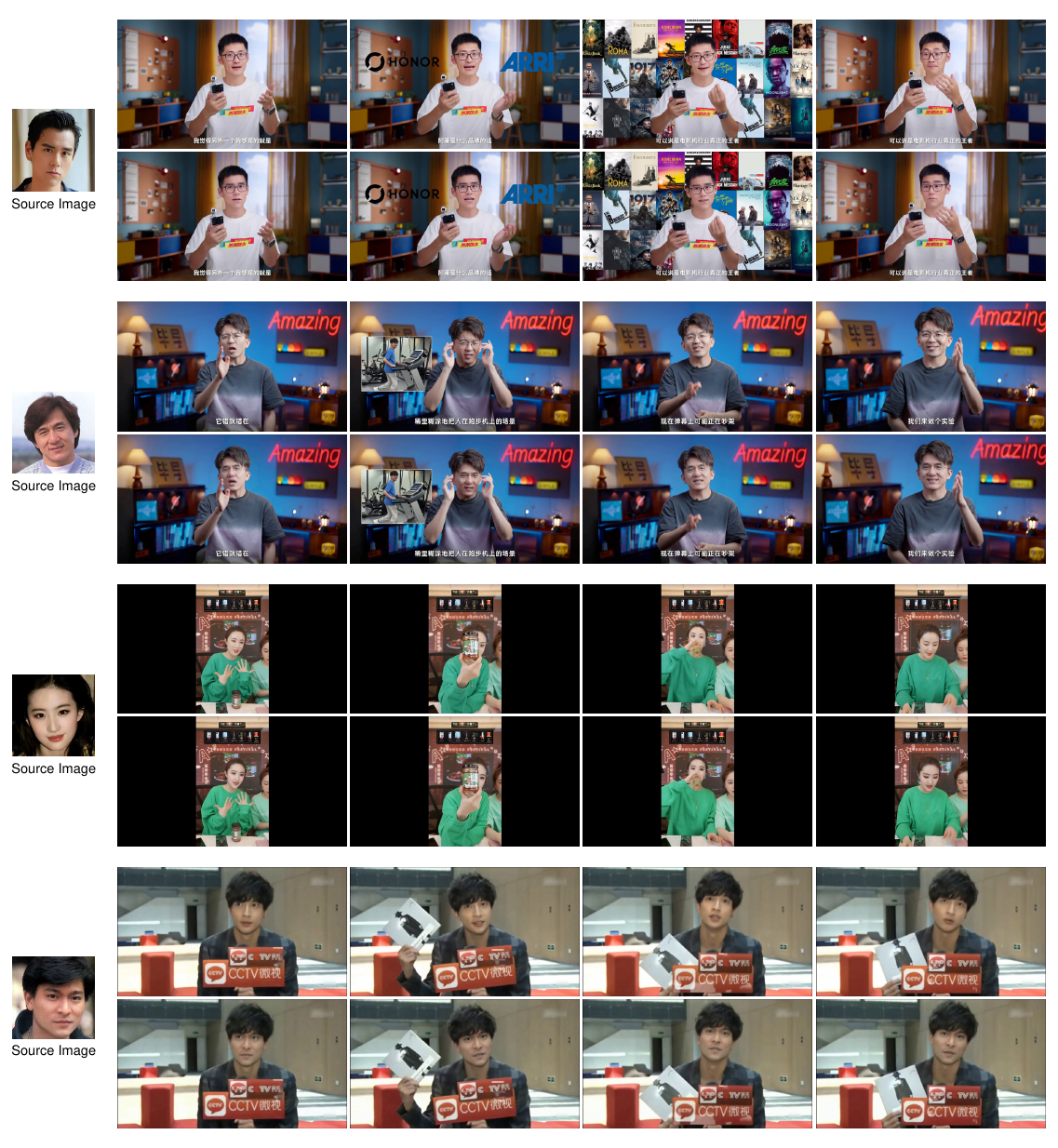}
    \caption{More qualitative results for live videos.}
    \label{fig:app_online}
\end{figure}

\begin{figure}[t]
    \centering
    \includegraphics[
        page=3,
        width=\textwidth,
        trim=25 55 25 55,
        clip
    ]{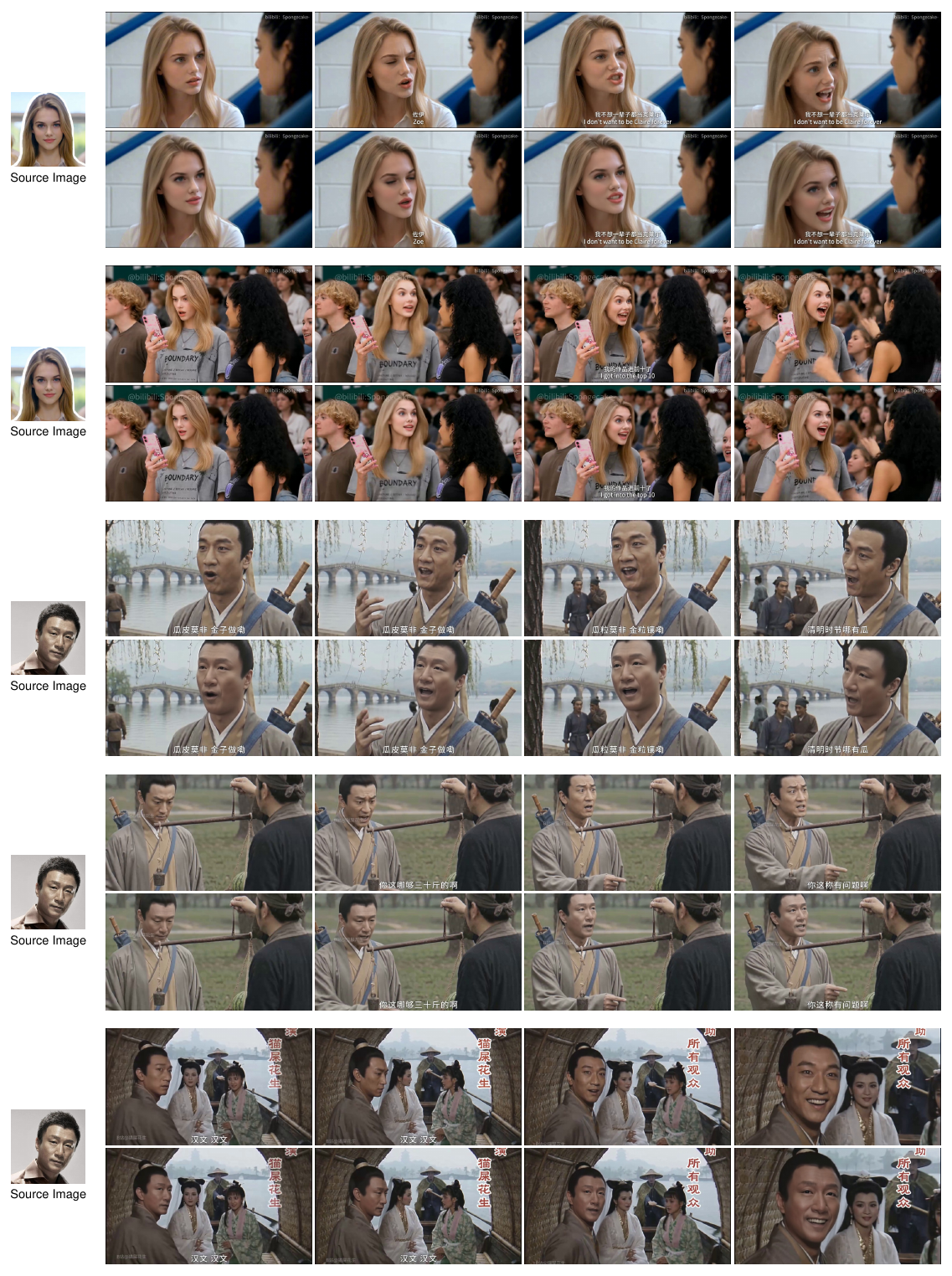}
    \caption{More qualitative results for AI-generated dramas.}
    \label{fig:app_drama}
\end{figure}

Figures~\ref{fig:app_movie}--\ref{fig:app_drama} illustrate applications of \modelname to movies, live and online videos, and AI-generated short dramas.

Movie footage requires identity editing to preserve both cinematic lighting and the nuances of the original performance.
The displayed results follow the target video's illumination, subtle facial expressions, and head poses, including shots with partial occlusion, integrating the source identity while retaining the scene's visual character.

Live and online videos require sustained streaming generation that follows a speaker's expressions and head movements.
Our method supports continuous processing with stable facial appearance in the displayed sequences, including partially occluded faces.
Its streaming throughput makes it suitable for live-video applications whose frame-rate and buffering requirements match the measured processing speed and latency.

AI-generated short dramas present another use case: reinforcing character identity when large motions, complex lighting, or exaggerated expressions cause identity drift.
By applying a consistent source identity to the generated footage, \modelname provides an efficient post-processing step that strengthens identity consistency while retaining the original performance and scene context, without regenerating the entire scene.

  \clearpage

\section{Experimental Settings}
\label{sec:exp_settings}

\textbf{Metrics.}
We evaluate identity transfer, target-attribute preservation, video quality, and inference efficiency.
Following DreamID-V~\citep{guo2026dreamid}, we measure identity transfer using cosine similarity between ArcFace~\citep{deng2019arcface} embeddings of the source image and generated frames, reported as ID Sim.
We also annotate the displayed outputs with frame-level ID Sim. scores to facilitate comparison between embedding similarity and visual appearance.
Following LivingSwap~\citep{luo2026preserving}, we assess target-attribute preservation through distances between generated and corresponding target frames in estimated head pose (Pose)~\citep{ruiz2018fine}, expression coefficients (Expr.)~\citep{deng2019accurate}, and lighting coefficients (Light)~\citep{deng2019accurate}.
Gaze preservation is measured by cosine similarity between their estimated 3D gaze vectors (Gaze)~\citep{abdelrahman2023l2cs}.
We evaluate video quality using Fréchet Video Distance (FVD).
For efficiency, FPS measures the number of newly generated video frames output per second by the complete pipeline, excluding reused history frames.
Latency measures the end-to-end time required for a single inference run.
Both measurements include VAE encoding, denoising, VAE decoding, and output compositing.

\textbf{Implementation Details.}
We initialize our model from the pretrained DreamID-V checkpoint~\citep{guo2026dreamid} and train Phase I and Phase II for 1,000 iterations each.
Both phases use a learning rate of $10^{-5}$ and a batch size of 8.
In Phase II, each training rollout consists of five autoregressive generation windows.
Training uses four NVIDIA H100 GPUs, while all inference-speed measurements are conducted on a single NVIDIA GeForce RTX 4090 GPU.
For inference acceleration, we use SageAttention~\citep{zhang2024sageattention2} for DiT self-attention and cross-attention, and compile Transformer blocks individually using \texttt{torch.compile}.
We also replace the native Wan VAE with LightTAE for Wan 2.1~\citep{modeltc_lightvae} for lightweight encoding and decoding.
For fair efficiency comparisons, we apply the same acceleration strategies to all compatible methods.

\section{Ablation of Training Phase and Denoising Steps}
\label{sec:ab_phase}

\begin{table*}[t]
\centering
\caption{
Ablation of training phase and denoising steps.
DreamID-V is the initialization model.
Phase I uses bidirectional clip generation, while Phase II adds self-rollout training for streaming generation.
All model are evaluated on short videos bench
}
\label{tab:phase_steps_ablation}

\begingroup
\setlength{\tabcolsep}{6pt}
\renewcommand{\arraystretch}{1.08}

\resizebox{0.8\textwidth}{!}{%
\begin{tabular}{lccccccc}
\toprule
Model
& Steps
& ID Sim.$\uparrow$
& Pose$\downarrow$
& Expr.$\downarrow$
& Light$\downarrow$
& Gaze$\uparrow$
& FPS$\uparrow$ \\
\midrule

DreamID-V (Initialization)
& 16
& 0.445 & 8.504 & 2.695 & 0.427 & 0.798
& 1.815 \\

\midrule

\modelname (Phase I)
& 25
& 0.573 & \underline{5.723} & 2.526 & 0.266 & \underline{0.873}
& 6.702 \\

\modelname (Phase I)
& 4
& \textbf{0.583} & 6.086 & \textbf{2.476}
& \underline{0.260} & 0.869
& \textbf{30.280} \\

\midrule

\modelname (Phase II, Stream)
& 4
& \underline{0.577} & \textbf{5.354} & \underline{2.484}
& \textbf{0.242} & \textbf{0.881}
& \underline{21.174} \\

\bottomrule
\end{tabular}%
}
\endgroup
\end{table*}

\begin{figure*}[t]
    \centering
    \includegraphics[width=\linewidth]{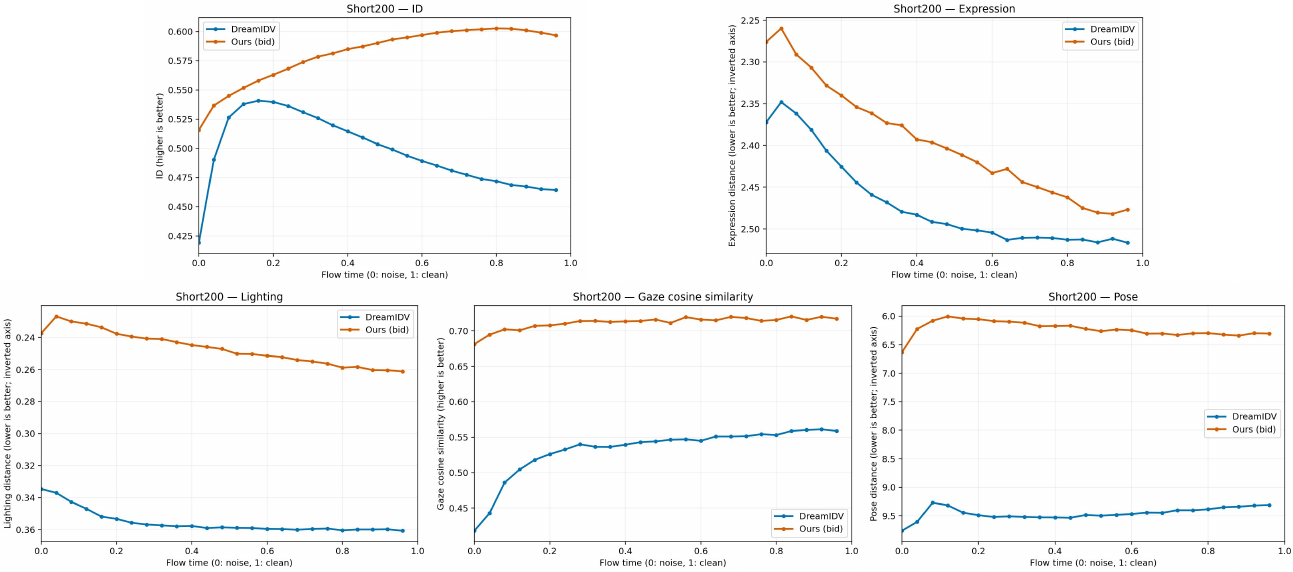}
    \vspace{-20pt}
    \caption{
    Identity and attribute metrics along the denoising trajectory on short200.
    Our bidirectional model and DreamID-V exhibit distinct trends across metrics.
    Flow time progresses from noise to clean data; error axes are inverted so that higher positions indicate better performance.
    }
    \vspace{-10pt}
    \label{fig:denoising_metrics}
\end{figure*}


\textbf{Setup.}
We compare the DreamID-V initialization with 16 denoising steps, the Phase-I bidirectional model with 25 and 4 steps, and the final Phase-II streaming model with 4 steps.
We evaluate identity similarity, target-attribute preservation, and end-to-end throughput to assess the effects of training phase and sampling budget.
To contextualize the sampling results, we also track identity and attribute metrics along the denoising trajectories of DreamID-V and the Phase-I model.



\textbf{Results and interpretation.}
As shown in Table~\ref{tab:phase_steps_ablation}, the four-step Phase-I model improves identity similarity from $0.445$ to $0.583$ over the initialization and improves all four target-attribute metrics.
Compared with 25-step sampling, it achieves comparable or slightly better identity, pose, expression, and lighting scores, together with higher gaze similarity, while increasing throughput from $6.702$ to $30.280$ FPS, a $4.52\times$ speedup.
The final streaming model retains most of this identity similarity ($0.577$) and further improves pose, lighting, and gaze scores, with a slight increase in expression error.
Its throughput decreases to $21.174$ FPS because streaming repeatedly processes overlapping history and incurs per-window overhead, while reused history frames are excluded from the output frame count.
These results support four-step sampling as an efficient choice for rollout training and streaming inference.

The trajectories in Fig.~\ref{fig:denoising_metrics} provide further context for the sampling results.
For our model, identity similarity improves through most of denoising before declining slightly near the endpoint, whereas expression and lighting errors generally increase after reaching their minima early in the trajectory.
Thus, progressing further along the denoising trajectory does not consistently improve the evaluated metrics.
One possible explanation is that late-stage refinement introduces appearance details that do not necessarily improve identity or attribute similarity as measured by the evaluation encoders.
This interpretation is consistent with the slightly better scores obtained with four-step sampling, although the trajectories alone do not establish why changing the step count produces these gains.
Four-step sampling may still sacrifice some fine visual detail relative to 25 steps, so its favorable identity and attribute scores should not be interpreted as uniformly better perceptual quality.

\section{Ablation of Joint Rec-ID Supervision}
\label{sec:ab_loss_weight}

\begin{figure*}[t]
    \centering
    \includegraphics[width=\linewidth]{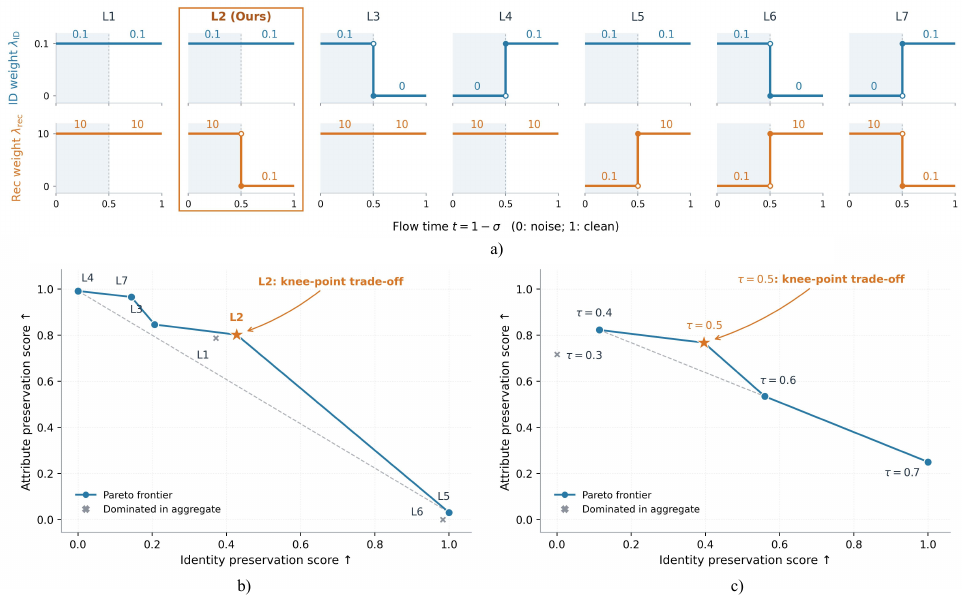}
    \vspace{-20pt}
    \caption{
    Joint Rec-ID supervision ablations on short200.
    (a) Weighting schedules.
    (b) Trade-offs across L1--L7.
    (c) Threshold sweep for L2.
    Scores average direction-aligned, min--max-normalized identity and attribute metrics within each sweep.
    Blue lines connect Pareto-optimal configurations; stars mark the endpoint-distance knee points.
    }
    \vspace{-10pt}
    \label{fig:loss_schedule}
\end{figure*}

\begin{figure*}[t]
    \centering
    \includegraphics[width=\linewidth]{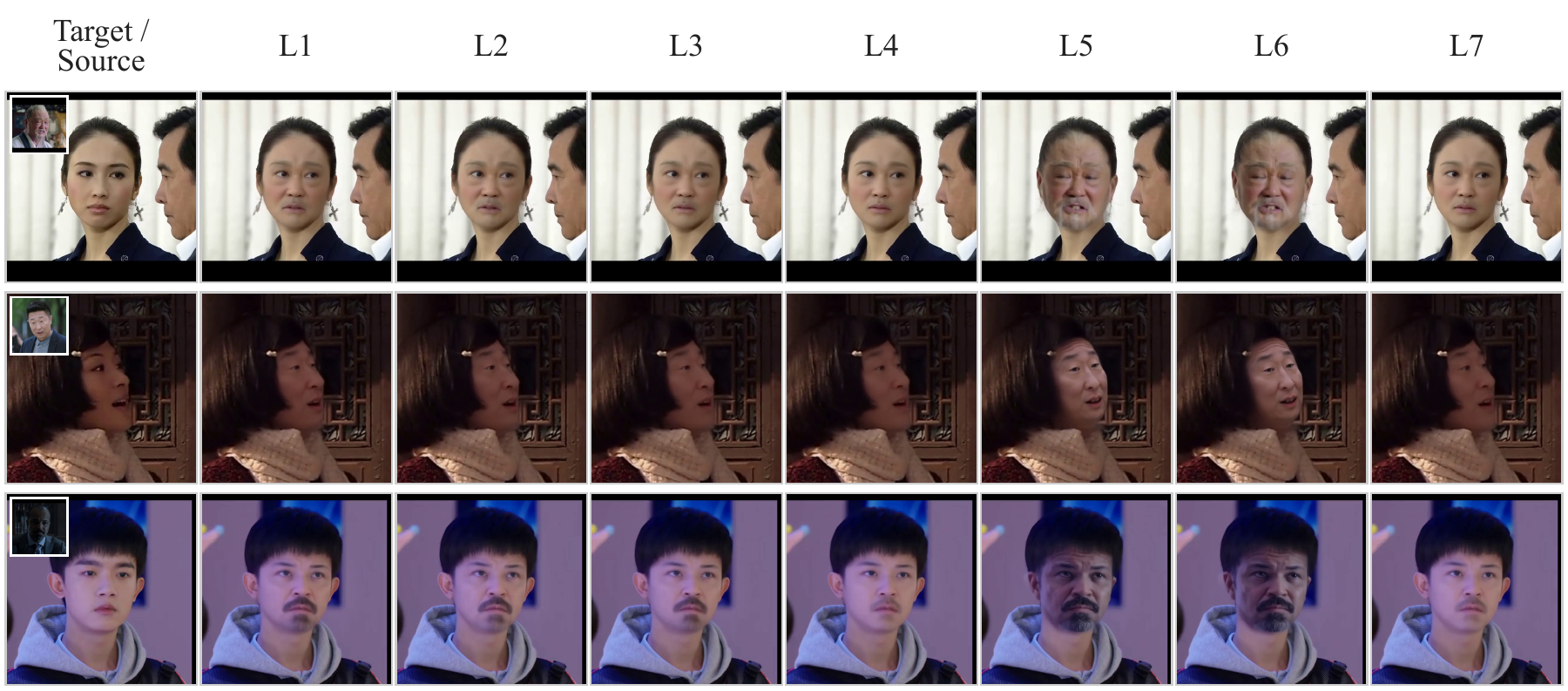}
    \vspace{-20pt}
    \caption{
    Qualitative comparison of joint reconstruction--identity supervision schedules (L1--L7).
    L2 balances source-identity transfer and target-attribute preservation.
    }
    \vspace{-10pt}
    \label{fig:loss_schedule_vis}
\end{figure*}

\begin{figure*}[t]
    \centering
    \begin{minipage}[c]{0.4\textwidth}
        \centering
        \includegraphics[width=\linewidth]{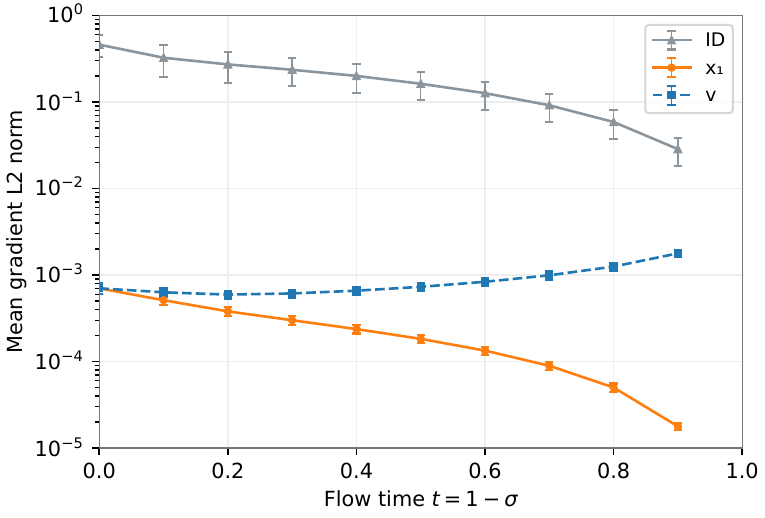}
        \vspace{-25pt}
        \caption{
            Gradient norms of identity supervision, reconstruction and velocity across noise levels.
        }
        \label{fig:endpoint_gradient}
    \end{minipage}\hfill
    \begin{minipage}[c]{0.6\textwidth}
        \centering
        \refstepcounter{table}
        \label{tab:endpoint_velocity}
         Table~\thetable: Reconstruction ablation on Short200.
        \medskip
        \scriptsize
        \setlength{\tabcolsep}{2pt}
        \renewcommand{\arraystretch}{1.3}
        \begin{tabular}{@{}lrrrrr@{}}
            \toprule
            Objective & ID Sim $\uparrow$ & Pose $\downarrow$ & Expr. $\downarrow$ & Light $\downarrow$ & Gaze $\uparrow$ \\
            \midrule
            $x_1$-pred & 0.573 & 5.723 & 2.526 & 0.266 & 0.873 \\
            $v$-pred & 0.569 & 5.655 & 2.534 & 0.278 & 0.865 \\
            \bottomrule
        \end{tabular}
    \end{minipage}
\end{figure*}

Our analysis in Sec.~\ref{sec:id_insight} motivates explicit identity supervision to complement latent reconstruction.
We examine how the weighting schedule, switching threshold, and reconstruction objective affect the balance between identity transfer and target-attribute preservation.

\textbf{Setup.}
We compare seven piecewise-constant weighting configurations, L1--L7, with the noise range divided at $\sigma=0.5$ (Fig.~\ref{fig:loss_schedule}(a)).
The identity weight is either $0.1$ or $0$, and the reconstruction weight is either $10$ or $0.1$.
For L2, we further vary the switching threshold $\tau$ over $\{0.3,0.4,0.5,0.6,0.7\}$, keeping the identity weight at $0.1$ and the high- and low-noise reconstruction weights at $10$ and $0.1$, respectively.
We also compare clean-endpoint and velocity reconstruction under the L2 schedule and examine the gradient norms of the identity and two reconstruction losses during training.
All configurations share the same initialization, training budget, and four-step evaluation protocol on the short200 benchmark of \datasetname.

\textbf{Weighting configurations.}
L2 balances identity transfer and target-attribute preservation by retaining identity supervision throughout training while reducing the reconstruction weight at low noise levels.
Compared with L1, it strengthens identity transfer with small changes in attribute preservation and imaging quality.
Removing identity supervision from either noise interval weakens identity transfer (L3/L4/L7), whereas weakening high-noise reconstruction compromises target fidelity despite stronger identity similarity (L5/L6).
The qualitative results in Fig.~\ref{fig:loss_schedule_vis} illustrate this trade-off, particularly the changes in expression and lighting under L5/L6.
Together with its knee-point position in Fig.~\ref{fig:loss_schedule}(b), these results support L2 as a balanced choice.

\textbf{Switching threshold.}
Increasing $\tau$ narrows the noise interval receiving strong reconstruction supervision, favoring identity transfer over target-attribute preservation.
Across the sweep, identity similarity increases alongside expression and lighting errors, while pose and gaze vary non-monotonically.
Under the normalized aggregation in Fig.~\ref{fig:loss_schedule}(c), $\tau=0.4$ achieves the highest attribute score and $\tau=0.7$ the highest identity score.
We use $\tau=0.5$, the knee point of this trade-off.


\textbf{Reconstruction objective.}
We compare clean-endpoint and velocity reconstruction while keeping the velocity-prediction architecture unchanged.
The training gradient norms in Fig.~\ref{fig:endpoint_gradient} indicate that endpoint reconstruction maintains a more consistent gradient scale relative to identity supervision across noise levels.
Under the L2 schedule, endpoint reconstruction achieves higher identity similarity, lower expression and lighting errors, and slightly higher gaze similarity, at the cost of higher pose error (Table~\ref{tab:endpoint_velocity}).
We therefore adopt endpoint reconstruction for its favorable balance between identity transfer and target-attribute preservation.

\section{Ablation of Streaming Window Size}
\label{sec:ab_window}

\begin{table*}[t]
\centering
\vspace{-10pt}
\caption{
Ablation of streaming window configurations.
$N_{\mathrm{ctx}}$ and $N_{\mathrm{gen}}$ denote the numbers of history and newly generated latent frames, respectively.
$^{\dagger}$ marks the default configuration. 
}
\label{tab:context_length_ablation}

\begingroup
\setlength{\tabcolsep}{6pt}
\renewcommand{\arraystretch}{1.08}

\resizebox{0.7\textwidth}{!}{%
\begin{tabular}{ccccccccc}
\toprule
$N_{\mathrm{ctx}}$
& $N_{\mathrm{gen}}$
& ID Sim.$\uparrow$
& Pose$\downarrow$
& Expr.$\downarrow$
& Light$\downarrow$
& Gaze$\uparrow$
& FPS$\uparrow$
& Latency (s)$\downarrow$ \\
\midrule

1 & 1
& 0.583 & 7.151 & 2.593 & 0.272 & 0.860
& 11.396 & \textbf{0.350} \\

1 & 3
& \underline{0.588} & 6.833 & 2.547 & 0.256 & 0.868
& 19.332 & \underline{0.592} \\

\midrule

3 & 3
& \textbf{0.592} & 6.780 & 2.562 & 0.256 & 0.867
& 16.063 & 0.848 \\

3$^{\dagger}$ & 9$^{\dagger}$
& 0.577 & 6.487 & \underline{2.488}
& \underline{0.239} & \underline{0.880}
& \textbf{21.174} & 1.782 \\

\midrule

5 & 9
& 0.576 & \underline{6.424} & \underline{2.488}
& 0.241 & 0.877
& 18.774 & 2.136 \\

5 & 12
& 0.568 & \textbf{6.292} & \textbf{2.453}
& \textbf{0.233} & \textbf{0.884}
& \underline{19.672} & 2.652 \\

\bottomrule
\multicolumn{9}{l}{\footnotesize
$^{\dagger}$Default configuration.}
\end{tabular}%
}
\vspace{-20pt}
\endgroup
\end{table*}

\textbf{Setup.}
We vary the history length $N_{\mathrm{ctx}}$ and generation length $N_{\mathrm{gen}}$ to examine the quality--efficiency trade-off in streaming inference.
Following Sec.~\ref{sec:phase_two}, each continuation generates $N_{\mathrm{gen}}$ new latent frames conditioned on $N_{\mathrm{ctx}}$ clean composited history frames, using bidirectional attention within the window.
Unlike the main results, which use a model trained with a fixed \((N_{\mathrm{ctx}}, N_{\mathrm{gen}})=(3,9)\) configuration, this ablation uses a model trained for 1,000 iterations with randomly sampled window sizes.
Table~\ref{tab:context_length_ablation} compares six configurations in terms of identity similarity, target-attribute preservation, and end-to-end throughput and latency.

\textbf{Results.}
Increasing $N_{\mathrm{gen}}$ at a fixed history length improves throughput but increases latency, whereas increasing $N_{\mathrm{ctx}}$ at a fixed generation length adds computational overhead and reduces throughput.
Among the evaluated configurations, $(N_{\mathrm{ctx}},N_{\mathrm{gen}})=(3,9)$ achieves the highest FPS with lower latency than the larger-window alternatives.
Target-attribute preservation generally improves with larger windows, while identity similarity peaks at $(3,3)$ and then declines.
We therefore adopt $(3,9)$ as a practical balance between throughput and latency, while maintaining competitive identity transfer and target-attribute preservation.

\section{Additional Qualitative Comparisons on Benchmarks}
\label{sec:more_vis}

\begin{figure*}[t]
    \centering
    \includegraphics[width=1.\linewidth]{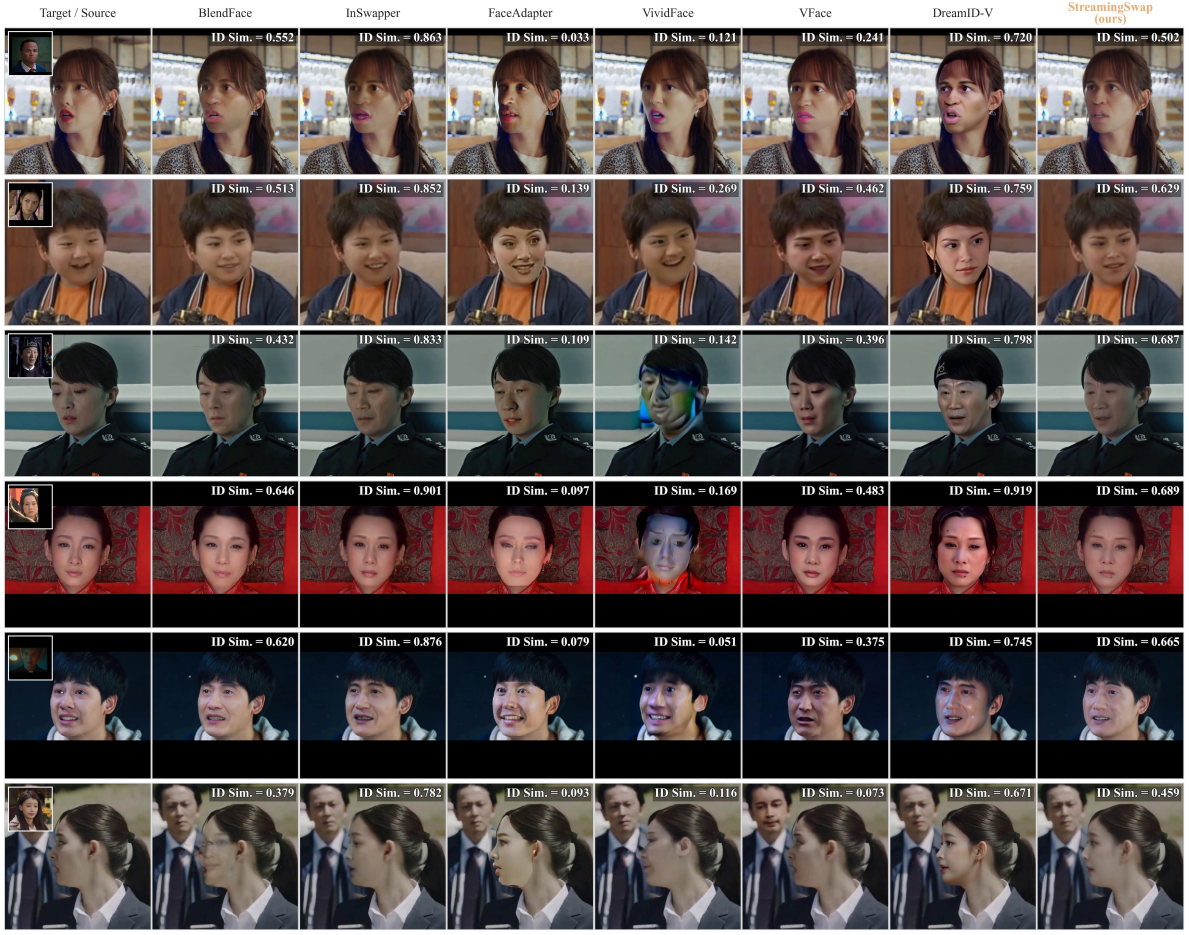}
    \vspace{-20pt}
    \caption{
        More qualitative results for method comparisons on short video bench.
    }
    \label{fig:compare2}
\end{figure*}

\begin{figure*}[h]
    \centering
    \includegraphics[width=1.\linewidth]{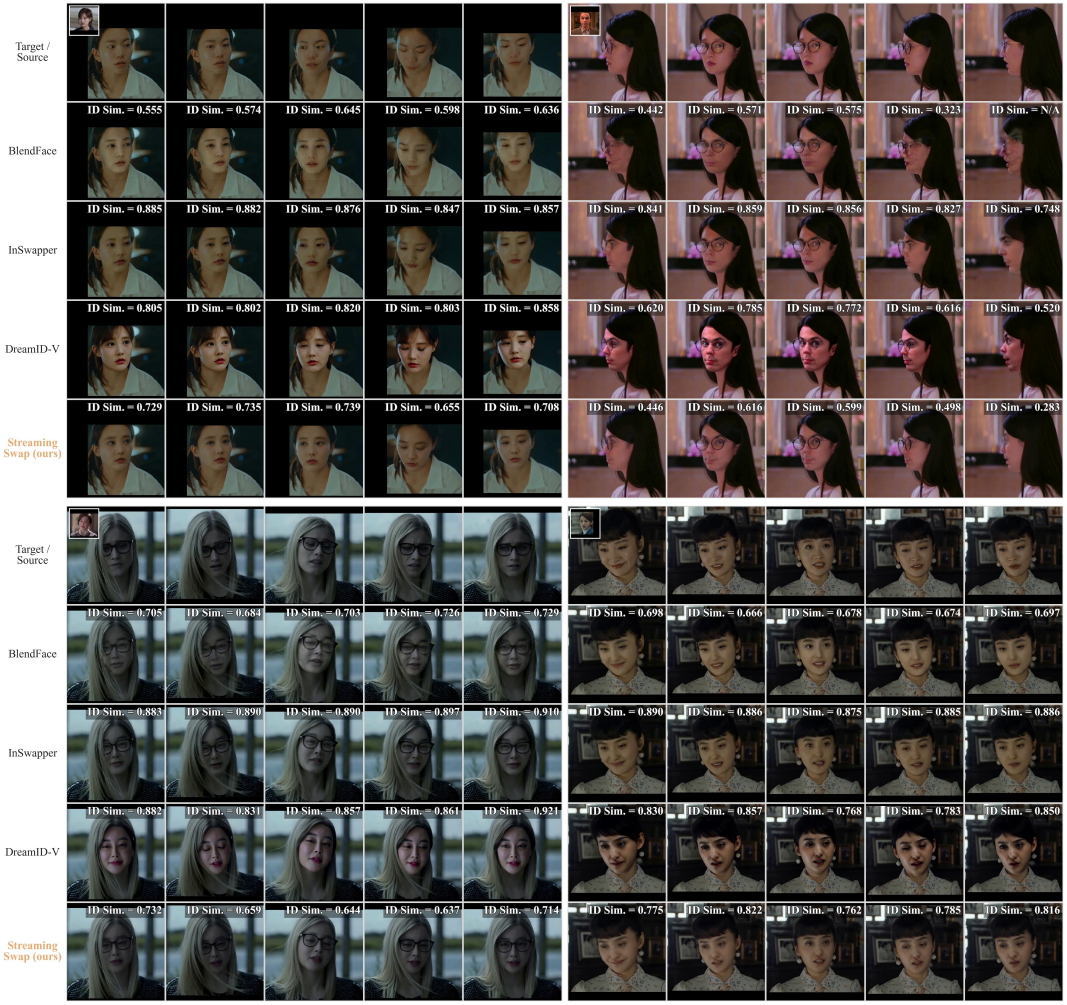}
    \vspace{-20pt}
    \caption{
        More qualitative results for method comparisons on long video bench.
    }
    \label{fig:compare3}
\end{figure*}

We provide additional qualitative comparisons in Fig.~\ref{fig:compare2},~\ref{fig:compare3} on the short and long video benchmarks to supplement the results presented in the main paper.
These examples further illustrate identity transfer and target-attribute preservation across diverse scenes.

\end{document}